%% file: main.tex
\input{Premable}

\usepackage{cleveref}
\usepackage{float}
\usepackage{amsmath} 
\usepackage{amssymb} 
\usepackage{booktabs}

\usepackage{hyperref}

\begin{document}
    \title{Aligning LLM agents with human learning and adjustment behavior: a dual agent approach}
	\author[1]{Tianming Liu}
    \author[2]{Jirong Yang}
	\author[1]{Yafeng Yin\footnote{Corresponding author. E-mail address: \textcolor{blue}{yafeng@umich.edu} (Y. Yin).}}
    \author[1]{Manzi Li}
    \author[3]{Linghao Wang}
    \author[3]{Zheng Zhu}

	\affil[1]{\small\emph{Department of Civil and Environmental Engineering, University of Michigan, Ann Arbor, United States}\normalsize}
 \affil[2]{\small\emph{Department of Electrical Engineering and Computer Science, University of Michigan, Ann Arbor, United States}\normalsize}
 \affil[3]{\small\emph{College of Civil Engineering and Architecture, Zhejiang University, Hangzhou, China}\normalsize}
	\date{October 31, 2025}
	\maketitle

\begin{abstract}
     Effective modeling of how human travelers learn and adjust their travel behavior from interacting with transportation systems is critical for system assessment and planning. However, this task is also difficult due to the complex cognition and decision-making involved in such behavior. Recent research has begun to leverage Large Language Model (LLM) agents for this task. Building on this, we introduce a novel dual-agent framework that enables continuous learning and alignment between LLM agents and human travelers on learning and adaptation behavior from online data streams. Our approach involves a set of LLM traveler agents, equipped with a memory system and a learnable persona, which serve as simulators for human travelers. To ensure behavioral alignment, we introduce an LLM calibration agent that leverages the reasoning and analytical capabilities of LLMs to train the personas of these traveler agents. Working together, this dual-agent system is designed to track and align the underlying decision-making mechanisms of travelers and produce realistic, adaptive simulations. Using a real-world dataset from a day-to-day route choice experiment, we show our approach significantly outperforms existing LLM-based methods in both individual behavioral alignment and aggregate simulation accuracy. Furthermore, we demonstrate that our method moves beyond simple behavioral mimicry to capture the evolution of underlying learning processes, a deeper alignment that fosters robust generalization. Overall, our framework provides a new approach for creating adaptive and behaviorally realistic agents to simulate travelers' learning and adaptation that can benefit transportation simulation and policy analysis.
     
\end{abstract}

\hfill\break%

\section{Introduction}
Agent-based modeling has become a pivotal paradigm in transportation system analysis and planning \citep{kagho2020agent}. This approach simulates system-level dynamics by modeling the interactions of autonomous agents, representing individual travelers, within a virtual environment. The performance of the underlying mobility infrastructure is determined by a traffic simulator in response to the agents' collective behavior. Crucially, this interactive process is designed to emulate human learning and adjustment: agents repeatedly engage with the infrastructure, receive performance feedback such as travel times and costs, and iteratively refine their travel plans and strategies. Therefore, the ability of agents to accurately model the learning and adjustment behavior of their human counterparts is critical for an agent-based model to correctly characterize transportation system dynamics and inform sound planning decisions. A mischaracterization of this behavior risks causing analysts to overlook emergent dynamics and can lead the simulation to converge on incorrect stationary points, thereby compromising the entire downstream planning process. Specifically, this can lead to misguided infrastructure investments based on inaccurate cost-benefit analyses or the rejection of effective policies due to a fundamental misunderstanding of public adaptation over time.

However, modeling the learning and adjustment behavior of travelers is a challenging task due to the complex cognitive process behind this behavior, beyond the impacts of the current system state and rationality. Indeed, research has shown that travelers' learning and decision-making are greatly impacted by their past travel experience and bounded rationality. On the impact of experience, research has shown that factors such as the utility of past travel experiences \citep{mahmassani2009learning,iida1992experimental,meneguzzer2013day}, travel time variability \citep{avineri2005sensitivity}, significant positive or negative experiences \citep{sonmez1998determining}, and exploration \citep{lowry2024multimodal} have significant impacts on how travelers make travel-related decisions, and it has also been shown that inclusion of past experiences can improve the modeling accuracy \citep{bamberg2003choice}. On the bounded rationality of travelers, existing studies have also found evidence of multiple types of bounded rational behavior in travelers' learning and adjustment process, including inertia \citep{qi2023investigating}, habit \citep{schlich2003habitual}, indifference between similar travel times \citep{zhao2016experiment}, and risk aversion/seeking \citep{avineri2005sensitivity,schwanen2009coping,zhang2018cumulative}. Overall, the complex interplay between traveler characteristics, past experience, and bounded rationality makes modeling human learning and adjustment challenging, especially for mathematical models that generally require \textit{a priori} assumptions. Furthermore, existing models such as day-to-day dynamic systems can simulate complex population-level outcomes but lack the resolution to analyze individual travelers, as they do not account for modeling individuals' decision-making processes \citep{jin2020stable,zhou2007structural,yu2020day}.

To address these limitations, researchers have recently proposed leveraging Large Language Models (LLMs) to advance agent-based modeling in transportation \citep{liu2025toward,adornetto2025generative,song2025incorporating,liu2025gatsim}. LLMs, such as GPT \citep{achiam2023gpt}, Gemini \citep{team2023gemini}, and Llama \citep{touvron2023llama}, are a class of generative artificial intelligence renowned for producing high-quality, human-like text from natural language instructions. This impressive general intelligence has prompted researchers to propose utilizing LLM agents, which are autonomous agents whose decisions are powered by LLMs, as the core of agent-based modeling frameworks. Compared to established models, LLMs offer distinct potential advantages. Their training on vast datasets of human-generated text, combined with their emergent capabilities for instruction-following \citep{wei2021finetuned} and zero-shot learning \citep{kojima2022large}, provides the potential to simulate complex human behaviors with fewer a priori assumptions \citep{mo2023large,liu2024can,liu2025toward,nie2025exploring}. Furthermore, their inherent ability to process and generate natural language simplifies the development and application of these agents, making them more accessible compared to complex mathematical models \citep{liu2025toward}.

Despite the potential of LLMs, realistically modeling the dynamic learning and adjustment processes of travelers remains a formidable challenge. Direct application of base LLMs is suboptimal, as recent studies have documented significant behavioral dissimilarities between them and human travelers \citep{wang2025comparing}. Consequently, methods are needed to condition LLMs into behaviorally plausible agents. However, research addressing this challenge is scarce, and existing methods invariably rely on prespecified behavioral assumptions \citep{wang2025agentic}. A clear research gap therefore exists for methods that can leverage the full expressive power of LLMs to create behaviorally accurate simulations of learning and adjustment, which our study aims to fill.

In this study, we propose a dual-LLM agent framework designed to automatically align the learning and adjustment behaviors of simulated agents with human travelers. At the center of our approach are two sets of LLM agents: (1) a series of LLM traveler agents, each of which simulates daily travel decisions based on a dynamic memory and a core persona, and (2) a LLM calibration agent, which leverages an LLM's reasoning to optimize the personas of the traveler agents against data streams of human travel behavior. Methodologically, our calibration agent is driven by a textual 'pseudo-gradient' descent for efficient refinement, and is further regularized by a smoothing mechanism to enhance generalization. Using a real-world dataset from a human day-to-day route choice experiment \citep{wang2025comparing}, we assess our approach and benchmark its performance against existing LLM agent baselines. The results show that our approach can achieve more accurate alignment of travelers' day-to-day learning behavior than existing approaches, as well as produce more accurate aggregate traffic flow simulation results. Furthermore, the evaluation results also reveal our method's ability to extend beyond replicating behavioral outcomes to aligning the underlying tendency and evolution of learning and decision-making. This deeper alignment enables our agent to capture how these strategies evolve over time, leading to robust generalization.

To our knowledge, our work presents the first framework for automatically calibrating LLM agents to simulate the dynamic learning and adjustment of travelers over time. Methodologically, we introduce a novel dual-agent architecture where an LLM calibration agent supervises the calibration of an LLM traveler simulation agent with online updates. This allows our system to capture the evolution of travel behavior and leverage the expressiveness of natural language to capture the complex behavior of human travelers. Practically, this framework enables researchers to leverage standard, API-based LLMs to create adaptive and realistic agents to integrate into agent-based simulations. Our approach contributes to the emerging literature on building more behaviorally realistic and longitudinally valid LLM-based tools for transportation modeling.

The remainder of this paper is structured as follows: \Cref{sec:related_work} provides a review of relevant literature on LLM traveler agents and existing methodologies of their alignment with human travelers, and also further positions our work. \Cref{sec:methodology} presents the mathematical settings of the problem and details our approach. \Cref{sec:empirical_setting} outlines our dataset, experiment setting, evaluation metrics, and compared approaches. The results and discussions of the experimental results are presented in \Cref{sec:result}. Finally, \Cref{sec:conclusion} concludes the paper.

\section{Related Work} \label{sec:related_work}
Our work is closely related to the recent stream of research on utilizing LLM for human travel behavior modeling and simulation. Below, we review the relevant existing works and further position the methodological contribution of our paper.

The advent of LLMs, with their sophisticated ability to interpret and generate natural language, has led researchers to explore their use as a ``silicon sample" of human behavior \citep{grossmann2023ai,hutson2023can}. In travel behavior research, LLMs are being developed to simulate travelers, offering a way to generate synthetic travelers and facilitate behavior and system modeling while relaxing the requirement of \textit{a priori} assumptions of traditional approaches \citep{liu2025toward,nie2025exploring}. Consequently, much literature has emerged to validate the behavioral alignment of base LLMs with human travelers across a wide spectrum of decisions. This includes extensive research into travel mode choice \citep{mo2023large,liu2024can,nishida2025large,xu2025evaluating}, destination choice \citep{wang2023would,beneduce2025large}, travel attitudes \citep{tzachristas2025guided,xu2025can}, the value of travel time \citep{yan2025valuing}, responses to policy changes \citep{yan2025addressing}, day-to-day learning \citep{wang2025comparing}, and parking search behavior \citep{fulman2025utilizing}.

Collectively, these studies confirm that LLMs can interpret travel scenarios and reason through decision-making processes in a human-like manner. They demonstrate an ability to leverage their internal knowledge base to generate plausible behaviors \citep{mo2023large,wang2023would,beneduce2025large,zhang2025transmode,yan2025addressing} and even replicate human tendencies and biases in various contexts \citep{yan2025valuing,fulman2025utilizing}. Despite these strengths, the literature also consistently identifies significant behavioral misalignments. Key limitations include discrepancies in preferences \citep{liu2024can,nishida2025large} and values \citep{xu2025can}, reduced sensitivity and elasticity to contextual factors \citep{yan2025valuing}, and divergent decision-making rules compared to humans \citep{wang2025comparing}. Overall, existing research forms a clear consensus: while LLMs hold immense potential for behavioral simulation, they require further conditioning to ensure their outputs are both accurate and reliable for research and application.

To address the challenges in conditioning LLMs for reliable behavioral simulation, recent research in travel behavior has increasingly focused on applying LLM agents, which are autonomous entities whose perception and decision-making processes are driven by an underlying LLM, to enhance alignment and facilitate behavioral simulation. Existing methodologies for developing these agents to accurately imitate human travel behavior primarily fall into two distinct categories,

\begin{itemize}
    \item \textbf{Memory system enhancement:} This approach equips LLM agents with a structured memory module to store past experiences (actions, outcomes, reflections). During decision simulation, relevant memories are retrieved and incorporated into the LLM's prompt, typically via retrieval-augmented generation techniques. This strategy is effective for two key reasons: (1) It mirrors the cognitive process where humans leverage past experiences for learning and decision-making. (2) It capitalizes on the LLM's few-shot learning capabilities \citep{brown2020language}, allowing the model to treat retrieved memories as contextual examples, discern underlying behavioral patterns, and apply them to new situations. The utility of memory-augmented LLMs for behavioral simulation has been demonstrated across various transportation contexts \citep{liu2024can,alsaleh2025towards,xu2025evaluating,zhang2025transmode,xu2025can}. Currently, techniques on using them to enhance LLM traveler agents include utilizing memory systems to improve multi-agent simulation convergence \citep{liu2025llm}, guide agent trajectory and activity planning through summarized experiences \citep{wang2024large,liu2025toward}, and enhance the simulation of specific choices like public transit usage or activity locations \citep{ge2025llm,chen2025perceptions}.

    \item \textbf{Persona learning:} This strategy assigns each LLM agent a persona that explicitly describes the intrinsic values, preferences, and decision-making characteristics of the human traveler being simulated. Analogous to parameters in traditional utility functions, this persona provides the LLM with a consistent identity to adopt during generation, thereby improving behavioral alignment. The primary methodology involves inferring these personas inversely from empirical behavioral data. Research has consistently shown that employing such inferred personas significantly enhances the fidelity of LLM agent simulations. For example, \cite{liu2024can} first applied preference personas that are inversely learned from a stated preference dataset to assist behavior simulation, and concluded that this approach can greatly enhance such alignment. Subsequent work has further validated the effectiveness of learning personas from various data sources to better simulate choices related to activity planning and mode selection \citep{chen2025perceptions,liu2025aligning,sameen2025synthesizing}.
\end{itemize}

To imitate human travelers' learning and adjustment behavior, it is necessary to provide LLM information on both the context (including past memories) and how the traveler makes decisions (included in the persona). Therefore, on aligning the learning and adjustment behavior, both the memory and the persona are necessary for the agents. In a closely related paper, \cite{wang2025agentic} proposed giving the agent a pre-assigned persona that describes the big-five personality traits \citep{roccas2002big} and a memory processing system of exponential smoothing to simulate travelers' day-to-day route choice behavior, and evaluated their alignment with human travelers. Compared to the aforementioned studies, our paper makes two main methodological contributions,
\begin{enumerate}
    \item \textbf{We relax the assumptions and restrictions on the agent persona.} Compared with existing studies that either employs \textit{a-priori} personas \citep{wang2025agentic} or a Likert scale measurement of pre-specified factors \citep{liu2024can,liu2025aligning,sameen2025synthesizing}, we relax the format restrictions of the pre-specified personas by reducing the formatting restrictions of persona descriptions, enabling the agents to better leverage natural language to describe complex human behavior.
    \item \textbf{To support this relaxation, we introduce automated learning guided by another LLM agent for the traveler agents.} Relaxation of the persona format leads to a large feasible region, thereby challenging the persona learning  procedure. Therefore, we introduce another LLM agent to automatically calibrate the agent persona, as LLM's language reasoning ability particularly suits this task. This idea is similar to recent efforts to leverage the ``LLM for Science" agents for automatic discovery in transportation \citep{lai2025llmlight,guo2025automating}, but ours is the first instance in using this approach to facilitate travel behavior alignment.
\end{enumerate}

\section{Methodology} \label{sec:methodology}

In this section, we formally define the LLM agent alignment problem and outline our approach. First, we present the definition of the dynamics of the transportation system and outline our approach in \Cref{subsec:method_basic}. The design and training process of the LLM traveler agents and the LLM calibration agent are then presented in \Cref{subsec:method_traveler_agent} and \Cref{subsec:method_train_agent}, respectively.

\subsection{Premise}
\label{subsec:method_basic}
\begin{figure}[H]
    \centering
    \includegraphics[width=0.8\linewidth]{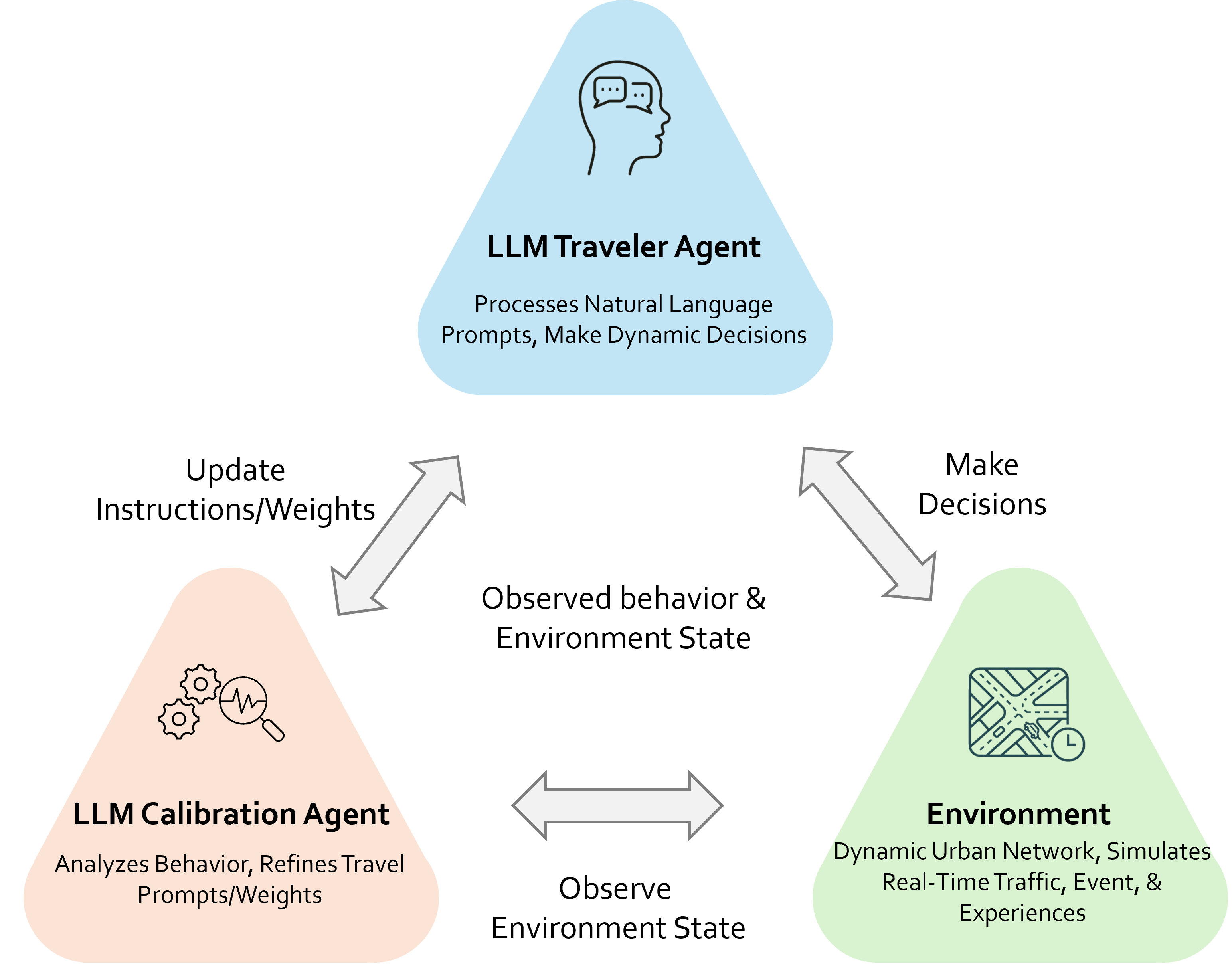}
    \caption{Overview of our approach}
    \label{fig:Overview}
\end{figure}

To establish our framework, we first define the transportation system and its dynamics. The system consists of a set of $M$ heterogeneous agents, $\mathcal{N}=\{n_1,n_2,\dots,n_M\}$, interacting with an environment, $\mathcal{E}$. Each agent $n_i \in \mathcal{N}$ represents a traveler, while the environment $\mathcal{E}$ encompasses the transportation infrastructure. The interaction between agents and the environment occurs over a sequence of discrete time periods, indexed by $t=1,2,\dots$. In each period $t$, the system evolves through the following sequential process,

\paragraph{1. Agent Decision-Making} In each period $t$, agent $n_i \in \mathcal{N}$ makes a travel decision, $a_i^t$. This decision is determined by the agent's decision-making function, $f_i$, which describes the agent's behavior of referring to their past experiences and making decisions based on their characteristics,
\begin{equation}
    a_i^t = f_i\left(\{a_i^\tau, o_i^\tau\}_{\tau=1}^{t-1} \right)
    \label{eq:agent_decision}
\end{equation}
\noindent in which $\{a_i^\tau, o_i^\tau\}_{\tau=1}^{t-1}$ is the collection of the agent's past decisions ($a_i^\tau$) and their observed outcomes ($o_i^\tau$).

\paragraph{2. Environment State Update} Following the agents' collective decisions $\{a_i^t\}_{i=1}^M$ in period $t$, the traffic condition changes and thus the environment transitions to a new state, $s^t$. This process is governed by a known system dynamics function, $E$, which updates the state based on the previous state $s^{t-1}$ and the agents' decisions,
\begin{equation}
    s^t = E\left(\{a_i^t\}_{i=1}^M, s^{t-1}\right)
    \label{eq:update_state}
\end{equation}
\noindent Here, the state vector $s^t$ describes the performance of the transportation infrastructure (e.g., link travel times, traffic volumes). The function $E$ represents the underlying model of the system, such as a set of link performance functions in a traffic assignment context. As is standard in agent-based modeling, the function $E$ is specified by the modeler to represent the ground-truth dynamics of the simulated environment, using tools such as traffic simulators.

\paragraph{3. Agent Observation} After the environment transitions to the new state $s^t$, each agent $n_i$ receives feedback based on their individual experience. Generally, in a transportation system, a traveler typically only knows a part of the system state, such as only knowing the travel time of their chosen route. To reflect this partial observability, we model the agent's personal observation, $o_i^t$, as a function of the global state and their specific decision,
\begin{equation}
    o_i^t = O(s^t, a_i^t)
    \label{eq:agent_perception}
\end{equation}
\noindent in which the observation function, $O$, maps the system state and the agent's decision to a personalized outcome. For example, $O$ would take the full set of network travel times ($s^t$) and the agent's chosen route ($a_i^t$) to return the specific travel time they experienced. This new experience pair, $(a_i^t, o_i^t)$, is then added to the agent's history, forming the basis for their decision in the next period.

As the interactive process defined by \Cref{eq:agent_decision,eq:update_state,eq:agent_perception} unfolds, the modeler continuously receives data on agent decisions, their observations, and the resulting system states. At each time period, the task for the modeler is to learn a set of surrogate decision-making models, $\{\hat{f}_i\}$, one for each agent. These surrogate models, $\hat{f}_i$, are designed to approximate the corresponding human traveler's true, unobserved behavioral strategies (represented by $f_i$ in \Cref{eq:agent_decision}). The goal is to use these learned models to generate simulations of the human agents' future actions ($a_i^t$) and the resulting aggregate system states. Therefore, the objective is to learn $\{\hat{f}_i\}$ so that each agent's future behavior is well-aligned with the behavior of the human it represents, and the modeler's goal can be conceptualized as finding a model $\hat{f}_i$ that minimizes the expected future behavior simulation error. Mathematically, at the end of time period $T$, the agent alignment optimization problem for the modeler is,
\begin{equation}
    \min_{\hat{f}_i} \mathbb{E}_{t > T} \left[ \rho\left(\hat{f}_i\left(\{a_i^\tau, o_i^\tau\}_{\tau=1}^{t-1}\right), a_i^t\right) \mid \{a_i^\tau, o_i^\tau\}_{\tau=1}^{T} \right]
    \label{eq:agent_learning_alignment}
\end{equation}
\noindent where the expectation $\mathbb{E}$ is taken over all possible future trajectories given the history from time period $1$ to $T$, and $\rho$ is a loss function that quantifies the discrepancy between the model's prediction and the corresponding traveler's actual decision. By solving \Cref{eq:agent_learning_alignment}, the modeler aims to create the most robust and generalizable behavior simulator of what the agent will do next that can be applied in an agent-based simulation.

To solve the optimization problem \ref{eq:agent_learning_alignment}, we propose a dual-LLM-agent approach that automatically learns the best-fitting decision-making function $f_i$ from the data. An overview of our approach is presented in \Cref{fig:Overview}.

\subsection{LLM traveler agents} \label{subsec:method_traveler_agent}
\begin{figure}[H]
    \centering
    \includegraphics[width=1\linewidth]{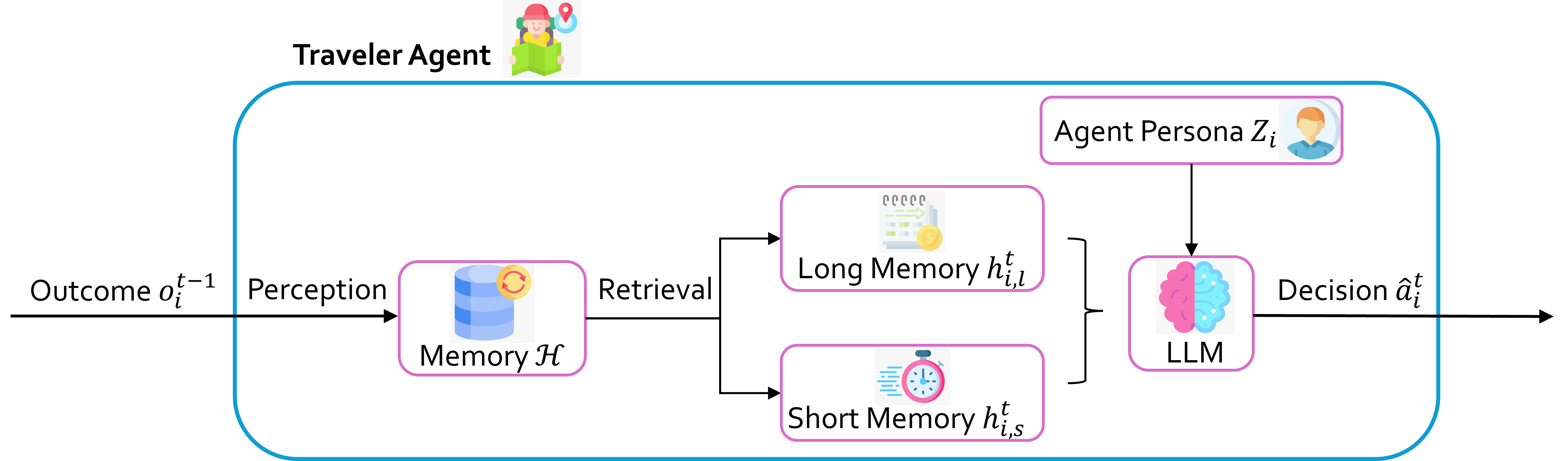}
    \caption{Components and workflow of an LLM traveler agent}
    \label{fig:Traveler_agent}
\end{figure}

First, we describe the design of the LLM traveler agent. This agent imitates the behavior of their human counterpart by simulating the human learning and decision-making processes defined in \Cref{eq:agent_decision} and \Cref{eq:agent_perception}. By specifying the workflow from perception to decision, the LLM agent provides a complete implementation of $\hat{f}_i$, and is capable of being calibrated and used in simulation. The architecture of this agent is illustrated in \Cref{fig:Traveler_agent}.

Each agent is composed of three core elements: the LLM core, the persona, and the memory. These elements are connected with each other in the agent workflow through the agent's core functionalities. Below, we first elaborate the core elements for an LLM agent that simulates the behavior of human traveler $i\in \mathcal{N}$:
\begin{enumerate}
    \item The LLM core (denoted by $LLM$): The LLM-core is the ``brain" of the agent. It integrates information and, more importantly, serves as the cognitive engine for simulating the human learning and decision-making process.
    \item The persona (denoted by $Z_i$): The persona is a natural language description of the agent's intrinsic characteristics and decision-making heuristics. It defines the agent's traits, tendencies, and factors it cares about in the decision-making process, and is a fundamental driver behind the agent's decision-making process.
    \item The memory (denoted by $\mathcal{H}_i$): The memory emulates a human storing past experiences as memories. It stores a record of the agent's past decisions and their observed outcomes.
\end{enumerate}

These elements are connected and operationalized in the workflow through three core functionalities: perception, memory retrieval (short-term and long-term), and decision-making. Next, we elaborate the working of each of them:
\begin{itemize}
    \item Perception: The perception functionality manages the updating of the agent's memory $\mathcal{H}_i$. At the end of each period $t$, the agent receives the outcome $o_i^t$ corresponding to the previous decision. It then incorporates this new experience into its memory by adding it to the memory set:
\begin{equation*}
    \mathcal{H}_i \leftarrow \mathcal{H}_i \cup \{(a_i^t, o_i^t)\}
\end{equation*}
By this update, the agent maintains a complete record of its experiences. For computational efficiency, the memory is stored as a series of numerical vectors.

\item Memory retrieval: When making a decision, the agent retrieves relevant experiences from its memory. To emulate the human tendency to rely on both recent events and long-term patterns \citep{baddeley1993recency,shohamy2015integrating,bornstein2017reminders}, we implement a dual-memory retrieval mechanism:
\begin{gather}
    h_{i,s}^t = G_{i,s}(\mathcal{H}_i,t)
    \label{eq:short_term_retrival} \\
    h_{i,l}^t = G_{i,l}(\mathcal{H}_i,t)
    \label{eq:long_term_retrival}
\end{gather}
\noindent in which $h_{i,s}^t$ and $h_{i,l}^t$ represent the retrieved short-term and long-term memories, respectively. The functions $G_{i,s}$ and $G_{i,l}$ are the corresponding time-invariant retrieval mechanisms. To ensure compatibility with the LLM, the retrieved numerical data is converted into natural language summaries by populating a structured textual template.

\item Decision-making: In the decision-making stage, the agent's persona and retrieved memories are integrated into a structured prompt, $q_i^t$:
\begin{equation}
    q_i^t = \text{Prompt}(Z_i, h_{i,s}^t, h_{i,l}^t)
    \label{eq:prompt}
\end{equation}
\noindent Following established methods \citep{liu2024can,liu2025aligning,sameen2025synthesizing}, this prompt synthesizes all influential factors into a coherent description that instructs the LLM on how to simulate the decision based on the persona, the past experiences, and the decision-making context. The LLM then processes this query to generate a simulated decision, denoted by $\hat{a}_i^t$:
\begin{equation}
\hat{a}_i^t = \text{LLM}(q_i^t)
\label{eq:LLM_decision}
\end{equation}
\noindent In this step, the LLM, guided by the prompt, simulates the agent's cognitive process to produce the decision $\hat{a}_i^t$.
\end{itemize}

\begin{algorithm}[h!]
\caption{LLM Traveler Agent Decision Process in Period $t$}
\label{alg:agent_decision_process}
\begin{algorithmic}[1]
\Require Agent's persona $Z_i$; Memory $\mathcal{H}_i$; Retrieval functions $G_{i,s}, G_{i,l}$; an LLM core.
\Ensure Simulated action $\hat{a}_i^t$.

\State \Comment{Retrieve relevant experiences from memory.}
\State $h_{i,s}^t \leftarrow G_{i,s}(\mathcal{H}_i, t)$ \Comment{Retrieve short-term memory.}
\State $h_{i,l}^t \leftarrow G_{i,l}(\mathcal{H}_i, t)$ \Comment{Retrieve long-term memory.}

\State \Comment{Construct the prompt for the LLM.}
\State $q_i^t \leftarrow \text{Prompt}(Z_i, h_{i,s}^t, h_{i,l}^t)$ \Comment{Integrate persona and memories into a prompt.}

\State \Comment{Generate the decision.}
\State $\hat{a}_i^t \leftarrow \text{LLM}(q_i^t)$ \Comment{Prompt the LLM to simulate the traveler's choice.}

\State \Return $\hat{a}_i^t$
\end{algorithmic}
\end{algorithm}

Overall, the sequential process for an agent making a decision in time period $t$ is presented in Algorithm \ref{alg:agent_decision_process}. In our framework, we designate the memory retrieval as a modeler-specified component, while the persona is the primary learnable element. This design choice is motivated by two principles. First, it focuses the calibration process on the primary source of unobserved heterogeneity: the agent's latent persona. While the memory contains a factual history of known events, the persona is the unknown variable that requires further calibration. Second, the cognitive process of interpreting and acting upon retrieved memories is fundamentally a function of an agent's intrinsic decision-making style, which is a part of what the persona is designed to capture. By specifying a retrieval heuristic based on domain knowledge, we delegate the complex task of contextualizing and weighting this historical information to the learned persona.

\subsection{LLM calibration agents} \label{subsec:method_train_agent}
Following the definition of the LLM traveler agents in \Cref{subsec:method_traveler_agent}, the agent behavior alignment problem in (\ref{eq:agent_learning_alignment}) can be transformed into calibrating a persona $Z_i$ for each agent to correspondingly minimize the expected future simulation error for that agent $i$:
\begin{equation}
\begin{aligned}
    &\min_{Z_i} &&\mathbb{E}_{t > T} \left[ \rho(\hat{a}_i^t, a_i^t) \mid \{a_i^\tau, o_i^\tau\}_{\tau=1}^{t} \right]\\
    &\text{s.t.} && \hat{a}_i^t = \text{LLM}(\text{Prompt}(Z_i, h_{i,s}^t, h_{i,l}^t))
\end{aligned}
\label{eq:agent_persona_training}
\end{equation}

\begin{figure}[H]
    \centering
    \includegraphics[width=1\linewidth]{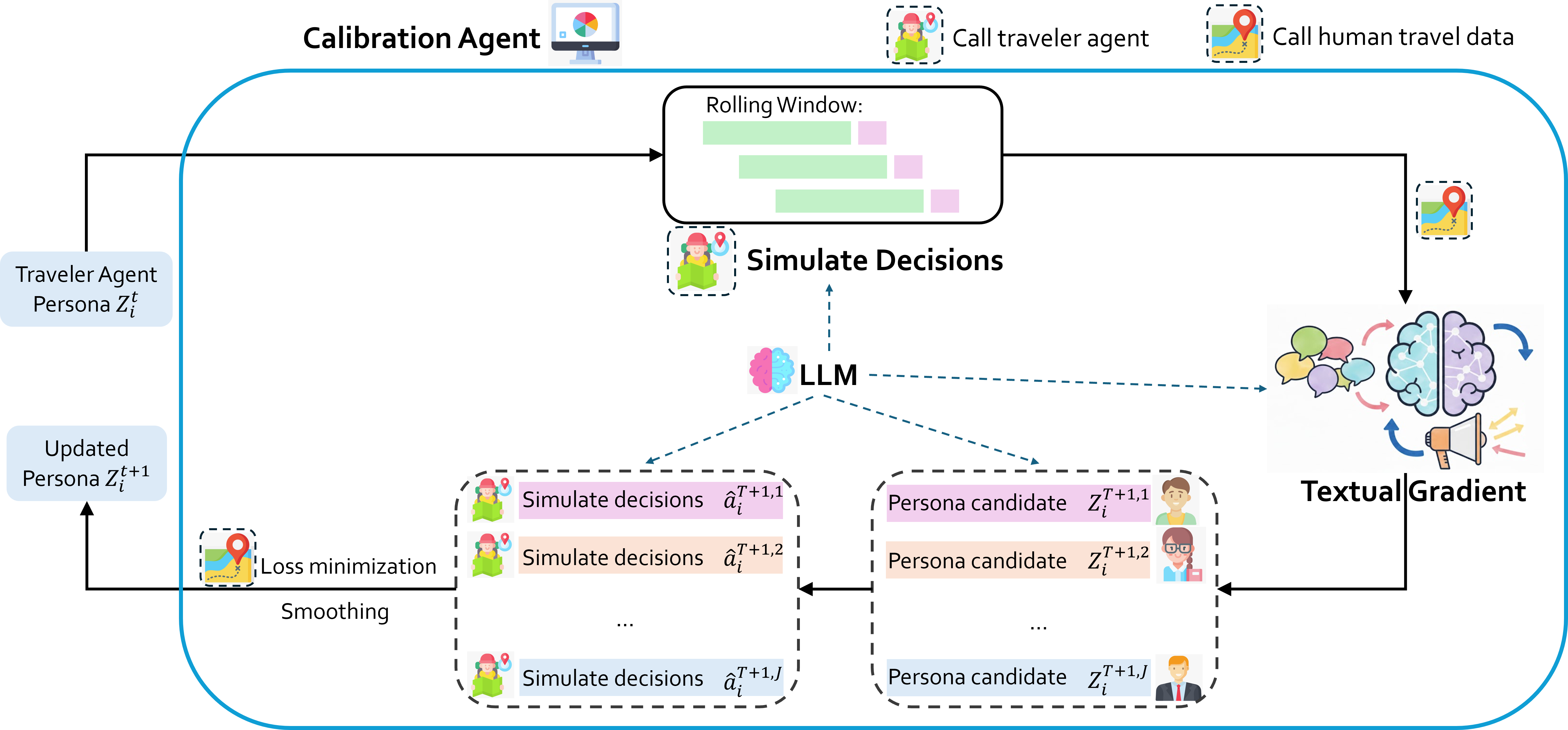}
    \caption{Components and workflow of the LLM calibration agent}
    \label{fig:Calibration_agent}
\end{figure}

In solving problem \ref{eq:agent_persona_training}, optimizing a free-form textual variable like $Z_i$ is challenging because the search space is high-dimensional, discrete, and non-differentiable. To address this, we build the LLM calibration agent that executes an automated calibration process. To address this challenge, we design an LLM calibration agent that executes an automated, iterative process. The core idea is to leverage the reasoning capabilities of an LLM to perform a textual ``pseudo-gradient" descent, inspired by similar methods in prompt optimization \citep{yuksekgonul2025optimizing}. The overall workflow of the automatic calibration agent is depicted in Figure \ref{fig:Calibration_agent}.

The calibration process is performed iteratively at the end of each time period $t$ when new observations of traveler behaviors and system state come to the modeler. To leverage the existing persona trained in the previous step, we do not retrain the agents' personas from scratch. Instead, we utilize a rolling update process to update the persona at every step based on the existing persona.

At the end of time $t$, the calibration agent focuses on a rolling window of the past $t_w$ periods, $[t-t_w+1, t]$, and uses data of human travelers' behavior during this window to conduct training. This approach aims to leverage the more recent observations of human travelers' behavior and prevent the number of training datapoints from growing with time at every step. Furthermore, focusing on more recent observations ensures the model adapts to the most recent thinking of the agent, which is critical for the persona to be up-to-date and able to generalize to the future. The calibration agent's workflow within each step $t$ proceeds as follows:

\paragraph{1. Evaluation of the Current Persona}
First, the calibration agent establishes a performance baseline for the current persona, $Z_i^t$. It simulates the agent's decisions over the active time window $[t-t_w+1, t]$ by calling the traveler agent's decision-making function $\text{LLM}$:
\begin{equation}
    \hat{a}_i^\tau = \text{LLM}\left(\text{Prompt}(Z_i^t, h_{i,s}^\tau, h_{i,l}^\tau)\right), \quad \forall \tau \in [t-t_w+1, t]
    \label{eq:LLM_step_original_persona_simulation}
\end{equation}

\paragraph{2. Pseudo-Gradient Generation}
Next, the agent generates a textual ``pseudo-gradient," $\frac{\Tilde{\partial}\rho}{\Tilde{\partial}Z_i^t}$, for each data point for agent $i$ in the active time window by analyzing the simulation results and attribute simulation errors to the persona. This is done by prompting the LLM core of the calibration agent (this agent functionality is denoted by $LLM_g$) to act as an analyst, identify the discrepancy between the simulated decision $\hat{a}_i^t$ and the human ground truth $a_i^t$, and suggest a textual correction to $Z_i^t$:
\begin{equation}
    \frac{\Tilde{\partial}\rho(\hat{a}_i^\tau, a_i^\tau)}{\Tilde{\partial}Z_i^t} = LLM_g(Z_i^t, \hat{a}_i^\tau, a_i^\tau), \quad \forall \tau \in [t-t_w+1, t]
\end{equation}
\noindent If the traveler agent correctly simulates the decision of the corresponding human traveler $(\hat{a}_i^\tau=a_i^\tau)$, the pseudo-gradient is null. If incorrect, it is a text describing the potential flaw in the persona that led to the error. We use individual data points to avoid overwhelming the LLM with long contexts, which is a known challenge for language models \citep{liu2023lost}.

\paragraph{3. Aggregation and Update Direction Synthesis}
The ``pseudo-gradients" of individual observations are then synthesized to form coherent update directions of the personas. To perform the synthesis, the LLM core of the calibration agent is instructed to process the set of ``pseudo-gradients" to identify common themes in their suggestion of editing of the personas, and propose $J$ distinct candidate update directions, $\{\Tilde{\nabla} Z_i^{t,j}\}_{j=1}^J$. This process is denoted by $LLM_i$, which is detailed below:
\begin{equation*}
    \Tilde{\nabla} Z_i^{t,j} = LLM_i\left(\left\{\frac{\Tilde{\partial}\rho(\hat{a}_i^\tau, a_i^\tau)}{\Tilde{\partial}Z_i^t}\right\}_{\tau=t-t_w+1}^{t}\right)
\end{equation*}
\noindent Instead of a simple mathematical aggregation of the individual pseudo-gradients, we employ a ``soft" aggregation method where an integrator LLM synthesizes the feedback. We use this ``soft" approach because the textual pseudo-gradients are heuristic error attributions, not true mathematical gradients, so their simple summation does not yield a valid global update direction. Furthermore, the search space of personas is a complex semantic domain. Therefore, to facilitate a more robust, multi-directional exploration of the textual search space at each iteration, we leverage the LLM to synthesize the collection of individual feedback points into multiple distinct and plausible improvement strategies.

\paragraph{4. Candidate Persona Generation}
Each update direction $\Tilde{\nabla} Z_i^{t,j}$ is then passed to the LLM core posing as a persona editor ($LLM_e$), which applies the suggested edits to the current persona $Z_i^t$ to generate a set of $J$ new candidate personas:
\begin{equation}
    Z_i^{t+1,j} = LLM_e(Z_i^t, \Tilde{\nabla}Z_i^{t,j})
\end{equation}

\paragraph{5. Candidate Evaluation and Selection}
Finally, the calibration agent evaluates all candidates. For each candidate persona $Z_i^{t+1,j}$, the calibration agent calls the traveler agent to simulate the decisions over the entire window to calculate its total loss, $\mathcal{L}_i^{t+1,j}$, and compare them against the loss of the original persona, $\mathcal{L}_i^t$. The losses are calculated as follows:
\begin{align*}
    &\mathcal{L}_i^{t+1,j} = \sum_{\tau=t-t_w+1}^{t} \rho(\hat{a}_i^{\tau, j}, a_i^\tau) \quad j=1,2,\dots,J\\
    &\mathcal{L}_i^t = \sum_{\tau=t-t_w+1}^{t} \rho(\hat{a}_i^{\tau}, a_i^\tau)
\end{align*}
\noindent in which $\hat{a}_i^{\tau, j}$ are the simulated decisions of the agents obtained by using the persona $Z_i^{t+1,j}$ in a similar fashion of \Cref{eq:LLM_step_original_persona_simulation}. After comparing the losses of candidates, the selected persona for the agent, $\tilde{Z}_i^{t+1}$, is the best-performing candidate. If an update is triggered, a smoothening step is performed to avoid significant overfitting of personas due to randomness in short-term observations. To conduct the smoothening, a long-term baseline persona is first synthesized by an LLM ($LLM_s$) from a curated set of recent human traveler observations of time window length $t_m$, $\{a_i^\tau,o_i^\tau\}_{t-t_s\leq \tau \leq t}$. This smoothing aims to mitigate potential significant fluctuations of personas due to a relatively short time window $t_w$ the agent is updated. Therefore, we set $t_m>t_w$ to use longer-term trends to smooth the training process. The final persona, $Z_i^{t+1}$, is then determined by prompting an integrator LLM ($LLM_{int}$) to merge the best candidate ($\tilde{Z}_i^{t+1}$) with long-term baseline. Overall, the update process is:

\begin{equation} 
Z_i^{t+1} = \begin{cases} 
LLM_{int}\left(\tilde{Z}_i^{t+1}, LLM_s\left(\{a_i^\tau,o_i^\tau\}_{t-t_m\leq \tau \leq t}\right)\right) & \text{if } \min_j \mathcal{L}_i^{t+1,j} < \mathcal{L}_i^t \\ 
Z_i^t & \text{otherwise} 
\end{cases} 
\end{equation} 

\noindent This integrated, conditional update ensures that the agent's persona is both responsive to recent feedback and robust against transient noise, leading to more plausible long-term behavior and a more stable training process. This final update of the persona completes one full step of the calibration process.

The overall workflow of the agent is summarized in \Cref{alg:calibration_agent_online}.

\begin{algorithm}[h!]
\caption{LLM Calibration Agent: Online Persona Optimization at Period $t$}
\label{alg:calibration_agent_online}
\begin{algorithmic}[1]
\Require Current persona $Z_i^t$; Human traveler data $\mathcal{D}_t = \{a_i^\tau, o_i^\tau\}_{\tau=t-t_s+1}^{t}$; Window sizes $t_w, t_m$ (with $t_m > t_w$); Candidate count $J$; LLM core.
\Ensure Updated persona for the next period, $Z_i^{t+1}$.

\State \Comment{\textbf{Step 1: Evaluate current persona.}}
\For{$\tau=t-t_w+1, \dots, t$}
    \State $\hat{a}_i^{\tau,t} \leftarrow LLM(\text{Prompt}(Z_i^t, h_{i,s}^\tau, h_{i,l}^\tau))$ \Comment{Simulate decisions with current persona.}
\EndFor
\State $\mathcal{L}_i^{t} \leftarrow \sum_{\tau=t-t_w+1}^{t}\rho(\hat{a}_i^{\tau,t}, a_i^\tau)$ \Comment{Calculate initial loss on $t_w$ window.}

\State \Comment{\textbf{Step 2: Generate pseudo-gradients.}}
\For{$\tau=t-t_w+1, \dots, t$}
    \State $\frac{\Tilde{\partial}\rho_\tau}{\Tilde{\partial}Z_i^t} \leftarrow LLM_g(Z_i^t, \hat{a}_i^{\tau,t}, a_i^\tau)$ \Comment{Generate textual feedback for each data point.}
\EndFor

\State \Comment{\textbf{Step 3: Synthesize update directions.}}
\For{$j=1, \dots, J$}
    \State $\Tilde{\nabla} Z_i^{t,j} \leftarrow LLM_i\Bigr(\{\frac{\Tilde{\partial}\rho_\tau}{\Tilde{\partial}Z_i^t}\}_{\tau=t-t_w+1}^{t}\Bigr)$ \Comment{Propose $j$-th update strategy.}
\EndFor

\State \Comment{\textbf{Step 4: Generate and evaluate candidate personas.}}
\For{$j=1, \dots, J$}
    \State $Z_i^{t+1,j} \leftarrow LLM_e(Z_i^t, \Tilde{\nabla}Z_i^{t,j})$ \Comment{Create $j$-th candidate persona.}
    \For{$\tau=t-t_w+1, \dots, t$}
        \State $\hat{a}_i^{\tau,t+1,j} \leftarrow LLM(\text{Prompt}(Z_i^{t+1,j}, h_{i,s}^\tau, h_{i,l}^\tau))$ \Comment{Re-simulate.}
    \EndFor
    \State $\mathcal{L}_i^{t+1,j} \leftarrow \sum_{\tau=t-t_w+1}^{t}\rho(\hat{a}_i^{\tau,t+1,j}, a_i^\tau)$ \Comment{Calculate candidate loss.}
\EndFor

\State \Comment{\textbf{Step 5: Select, smooth, and update persona.}}
\State $j^* \leftarrow \arg\min_j \mathcal{L}_i^{t+1,j}$
\If{$\mathcal{L}_i^{t+1,j^*} < \mathcal{L}_i^{t}$}
    \State $\tilde{Z}_i^{t+1} \leftarrow Z_i^{t+1,j^*}$ \Comment{Select best-performing candidate.}
    \State $Z_{base} \leftarrow LLM_s\left(\{a_i^\tau,o_i^\tau\}_{\tau=t-t_m+1}^{t}\right)$ \Comment{Synthesize long-term baseline.}
    \State $Z_i^{t+1} \leftarrow LLM_{int}(\tilde{Z}_i^{t+1}, Z_{base})$ \Comment{Merge best candidate with baseline.}
\Else
    \State $Z_i^{t+1} \leftarrow Z_i^t$ \Comment{Retain original persona.}
\EndIf

\State \Return $Z_i^{t+1}$
\end{algorithmic}
\end{algorithm}

\section{Empirical Experiment} \label{sec:empirical_setting}

\subsection{Experiment setting}
To train the LLM traveler agents and evaluate the performance of our proposed approach, we use the data from the real-world experiment conducted in \cite{wang2025comparing}. In the experiment, multiple groups of human participants are invited to participate in a day-to-day route choice experiment. In each experiment, a total of 15 participants are split into 2 OD groups. OD group 1 has 9 participants choosing between an ``expressway" and an arterial ``local 1", and the other 6 participants choosing between the same ``expressway" and a different arterial ``local 2". At the beginning of each round, each traveler is presented with the total number of people choosing the ``expressway" and the local arterial in the corresponding OD group, as well as their chosen route in the last round and that route's travel choice. After all participants have made their choice, the travel time of each route is determined by the BPR function:
\begin{equation*}
    t=t_0\Bigl(1+\beta(\frac{v}{c})^4\Bigr)
\end{equation*}

The specific route travel time function parameters are set up as follows:
\begin{itemize}
    \item Expressway: $t_0=5$, $\beta=0.075$, $c=3$.
    \item Arterials ``local1" and ``local2": $t_0=15$, $\beta=0.15$, $c=5$.
\end{itemize}
\noindent in this setting, the expressive has a shorter free-flow travel time but has a smaller capacity and more susceptible to congestion. In contrast, the local arterial has a longer free-flow travel time but has a higher capacity and impacted less by severe congestion. 

The route choice game is played for 160 rounds. We select the first experiment group for testing of evaluation of our approach, and use the first 80 rounds for training the agent and the last 80 rounds for evaluation. The route choice share of each human in the selected dataset is shown in \Cref{tab:human_mode_share}. For a more detailed description of the dataset, readers can refer to \cite{wang2025comparing} and its corresponding supplemental materials.

\begin{table}[h!]
	\caption{Behavior of each agent}\label{tab:human_mode_share}
	\begin{center}
		\begin{tabular}{l c c c c  c }
        \hline
        \multirow{2}{*}{OD group} & 
        \multirow{2}{*}{ID} & \multicolumn{2}{c}{First 20 days} & \multicolumn{2}{c}{Last 20 days}\\
        \cline{3-4} \cline{5-6}
        & & Local  & Expressway  & Local  & Expressway  \\
        \hline
        \multirow{9}{*}{Group 1}& 1 & 37.5\% & 62.5\% & 32.5\% & 67.5\%\\ 
        & 2 & 62.5\% & 37.5\% & 42.5\% & 57.5\%\\
        & 3 & 40.0\% & 60.0\% & 37.5\% & 62.5\%\\
        & 4 & 87.5\% & 12.5\% & 90.0\% & 10.0\%\\
        & 5 & 38.7\% & 61.3\% & 68.7\% & 31.3\%\\
        & 6 & 76.3\% & 23.7\% & 66.2\% & 33.7\%\\
        & 7 & 21.3\% & 78.7\% & 28.7\% & 71.3\%\\
        & 8 & 26.3\% & 73.7\% & 52.5\% & 47.5\%\\
        & 9 & 48.7\% & 51.3\% & 23.7\% & 76.3\%\\ 
        \hdashline
        \multirow{6}{*}{Group 2} & 10 & 70.0\% & 30.0\% & 92.5\% & 7.5\%\\
        & 11 & 95.0\% & 5.0\% & 98.7\% & 1.3\%\\
        & 12 & 62.5\% & 37.5\% & 67.5\% & 32.5\%\\
        & 13 & 36.3\% & 63.7\% & 55.0\% & 45.0\%\\
        & 14 & 90.0\% & 10.0\% & 40.0\% & 60.0\%\\
        & 15 & 35.0\% & 65.0\% & 62.5\% & 37.5\%\\
\hline
		\end{tabular}
	\end{center}
\end{table}\par

As detailed in \Cref{tab:human_mode_share}, the dataset is characterized by a rich heterogeneity in decision-making strategies. We observe a range of behaviors, from participants who maintain a consistent balance between routes (e.g., agents 1, 3, 12) to those with strong, persistent tendencies (e.g., agents 4 and 11 favor the arterial, while agent 7 favors the expressway). Crucially, the experimental design also captures non-stationary behavior, revealing several participants (e.g., agents 2, 5, 8, 13, 14, 15) who substantially alter their choice patterns over time. This diversity in both preferences and strategy evolution provides an ideal testbed for assessing a model's ability to learn proper underlying personas that can generalize to unseen conditions.

\subsection{Agent specification}

First, we specify the long-term and short-term memory retrieval as retrieving past memories of a certain length:
\begin{align*}
    &G_{i,s}(\mathcal{H}_i,t)=\{a_i^\tau,o_i^\tau\}_{\tau=\max(1,t-t_s)}^{t-1}\\
    &G_{i,l}(\mathcal{H}_i,t)=\{a_i^\tau,o_i^\tau\}_{\tau=\max(1,t-t_l)}^{t-1}
\end{align*}

\noindent in which $t_s$ and $t_l$ are lengths of the recent time periods contained in the short and long-term memories, respectively. Therefore, with the indexing of memories, each retrieval task can be done relatively easily by regular programming. In the experiment, we set $t_s=4$ and $t_l=24$.

We use the GPT-4o API as the LLM core for both the LLM traveler agents and the LLM calibration agent. For the training process, we use a rolling window of length 8 ($t_w=8)$ at each evaluation step, and use a context size of ($t_m=80$) for smoothing.

\subsection{Comparative metrics}

To compare the individual behavior alignment between the LLM traveler agent and the human traveler, we use the following metrics to quantify such a difference,
\begin{itemize}
    \item \textbf{Accuracy}: We use the average prediction accuracy of each method in correctly predicting the human's decision, controlling for the context and past experience:
    \begin{equation*}
    Accuracy=\frac{\sum_{i=1}^{15}\sum_{t=81}^{160}\mathbb{1}(\hat{a}_i^t=a_i^t)}{15\times 80}
    \end{equation*}

    \item \textbf{Weighted-Average F1 Score}: Since the humans in the experiment are split into two groups, each facing a different choice set for their decision at all times, we use the weighted-average F1-score to quantify the overall simulation performance. The weighted-average F1-score is calculated by:
    \begin{equation*}
        F1=\frac{3}{5}F1_1+\frac{2}{5}F1_2
    \end{equation*}
    \noindent in which $F1_k$ is the F1-score calculated for OD-group $k$, calculated as:
    \begin{equation*}
        F1_k=\frac{2\times Precision_k\times Recall_k}{Precision_k+Recall_k}
    \end{equation*}
    Overall, the F1-score is calculated as the population-weighted average of the F1-score for each group, which quantifies the binary choice simulation accuracy for each OD pair in the experiment. The resulting weighted average F1-score is a value between 0 and 1, with a higher F1-score represents higher simulation performance.
    \item \textbf{Behavior Vector}: To evaluate the alignment between simulated agents and human participants not only in decision outcomes but also in their \emph{learning tendencies}, 
we adopt the classification metric proposed by \cite{wang2025comparing}. 
For each traveler $i$ at day $t$, let $a_i^t$ denote the route they choose, 
$t_i^t$ denote the corresponding experienced travel time, 
and $\Tilde{t}_i^t$ the average travel time of the \emph{alternative} route in the same origin--destination (OD) group. 
Based on whether the traveler’s last trip was better or worse than the alternative and whether they switch routes, 
each daily decision can be categorized into four types:
\begin{equation}
    type_i^t=\left\{
    \begin{array}{ll}
    C^- &:~ t_i^{t-1}>\Tilde{t}_i^{t-1}\ \text{and}\ a_i^{t-1}\neq a_i^t \quad\text{(switch after loss)}\\
    S^+ &:~ t_i^{t-1}>\Tilde{t}_i^{t-1}\ \text{and}\ a_i^{t-1}=a_i^t \quad\text{(stay after loss)}\\
    C^+ &:~ t_i^{t-1}\leq\Tilde{t}_i^{t-1}\ \text{and}\ a_i^{t-1}\neq a_i^t \quad\text{(switch after win)}\\
    S^- &:~ t_i^{t-1}\leq\Tilde{t}_i^{t-1}\ \text{and}\ a_i^{t-1}=a_i^t \quad\text{(stay after win)}\\
    \end{array}
    \right.
\end{equation}
\noindent
These four cases respectively represent a traveler's change or stay decision following a loss or win experience.
Combining observations of agent behavior on each day, each agent's overall adjustment tendency over the entire experiment is summarized by a two-dimensional behavior vector:
\begin{equation}
\mathbf{v}_i = (C_i^-,\, S_i^-)
\end{equation}
where $C_i^-$ is the empirical probability of ``switching after loss,'' 
and $S_i^-$ is the probability of ``staying after win.'' 
This vector captures both the traveler's responsiveness to negative feedback and their inertia after positive reinforcement. Based on this representation, travelers can be categorized into four behavioral archetypes: 
\textit{naive} $(1,1)$, \textit{strategic} $(0,0)$, \textit{exploratory} $(1,0)$, and \textit{status quo} $(0,1)$.
While \cite{wang2025comparing} used these discrete types for classification, 
we adopt a finer-grained approach by directly comparing the continuous behavior vectors of humans and LLM agents 
through cosine similarity:
\begin{equation}
\text{Sim}_{\text{behavior}} =
\frac{\bar{\mathbf{v}}_{\text{LLM}}\!\cdot\!\bar{\mathbf{v}}_{\text{Human}}}
{\|\bar{\mathbf{v}}_{\text{LLM}}\|_2\times\|\bar{\mathbf{v}}_{\text{Human}}\|_2}.
\end{equation}
\noindent in which $||\cdot||_2$ is the $L^2$ norm. A higher $\text{Sim}_{\text{behavior}}$ indicates closer alignment in learning and adjustment patterns.
\end{itemize}




To quantify the route flow progression difference, we use the mean average percentage error (MAPE) and mean squared error (MSE) between the actual and simulated route flows. The MAPE is calculated by:
\begin{equation*}
    MAPE=\frac{1}{3\times 80}\sum_{t=81}^{160}||\frac{\hat{s}^t-s^t}{s^t}||_1
\end{equation*}
\noindent in which $\hat{s}^t,s^t \in \mathbb{R}^3$ are the simulated route flow and the ground-truth route flow after time period $t$, correspondingly, and $||\cdot||_1$ is the $L^1$ norm. The MAPE quantifies the average simulation error of route flow over the routes for the testing period.

The MSE is quantified by:
\begin{equation*}
    MSE=\frac{1}{3\times 80}\sum_{t=81}^{160}||\hat{s}^t-s^t||_2
\end{equation*}
\noindent in which $||\cdot||_2$ is the $L^2$ norm. The MAPE and MSE reflect slightly different characteristics, as the latter penalizes significant deviations of the simulation from the ground truth more. Therefore, we include both metrics in our evaluation.

\subsection{Compared methods}
We compare our proposed approach with existing LLM-based methods and other theory-driven methods in terms of performance. The baseline methods are each described below:
\begin{itemize}
    \item Base-LLM \citep{wang2025comparing}: In this approach, the LLM agent retains a memory of past travel experiences, but the travel decisions are made by an unmodified LLM. At each decision-making point, the LLM is prompted with the travel context and past travel experiences before making a travel decision. The specific prompts are consistent with those of \cite{wang2025comparing}.

    \item Recursive-LLM \citep{wang2025agentic}: In this approach, the LLM agent not only retains memory of past experiences, but also process and update their perceptions of route travel times via a recursive exponential smoothing process. For each travel decision, the LLM agents refers to their perception of route travel times as well as how many time have each route has been explored to make decisions. It uses the established day-to-day simulation paradigm but replaces the traditional agent decision-making module with a more flexible LLM agent. The specific application prompts in our assessment are consistent with those described in \cite{wang2025agentic}.

    \item Bounded-LLM: In this approach, we condition the agents with the learning model described by \cite{jotisankasa2006framework}. The learning and updating of perceived travel time is only activated by a bounded rationality trigger; travelers only update their beliefs if their actual experienced travel time is surprisingly different from their perceived time, exceeding a personal threshold. The agent's route choice in the next time period is then made using a random utility maximization model, which is heavily influenced by habit; if the learning trigger was not activated, travelers are highly likely to repeat their previous choice.
\end{itemize}

\section{Results} \label{sec:result}
\subsection{Comparison of individual-level choice behavior}
First, we evaluate the accuracy of individual behavior predictions by comparing our proposed method to several baselines. \Cref{tab:ind_pred_results} reports the average accuracy and F1-score, along with the win rates of our method versus each baseline over the individual agents.
\begin{table}[h!]
	\caption{Comparison of our model with established models on individual behavior prediction performance}\label{tab:ind_pred_results}
	\begin{center}
		\begin{tabular}{l c c c c}
        \hline
        & \multicolumn{2}{c}{Performance metrics} & \multicolumn{2}{c}{Win rate of our method on individuals} \\
        \cline{2-3} \cline{4-5}
         & Accuracy & F1-score &  On accuracy & On F1-score\\
        \hline
        Base-LLM & 0.549 & 0.544 & 0.800 & 0.733\\
        Recursive-LLM & 0.413 & 0.372 & 0.666 & 0.733\\
        Bounded-LLM & 0.411 & 0.374 & 0.800 & 0.800\\
        Ours &  \textbf{0.655} & \textbf{0.635}  & - & - \\
\hline
		\end{tabular}
	\end{center}
\end{table}\par

In the results shown in \Cref{tab:ind_pred_results}, the proposed method shows a significant performance advantage over existing approaches across both aggregate and individual metrics. At the aggregate level, our approach yields a relative improvement of 19.3\% in accuracy (from 0.549 to 0.655) and 16.7\% in F1-score (from 0.544 to 0.635) over the best-performing baseline. This superiority is also evident at the individual level, as our method outperforms all baselines for at least 66.6\% (10/15) of agents. This steady gain on both levels clearly demonstrates the superior performance of our agent. Interestingly, the theory-driven LLM agents underperformed in our experiment, as they both score below the LLM baseline with only a memory in \cite{wang2025comparing}. We posit that this is because the participants' learning and adjustment behaviors deviated substantially from the perception update and bounded rational rules specified in classic theoretical models, thereby causing them to perform less effectively. This comparison also demonstrates the limitation in posing behavioral models with \textit{a priori} behavior assumptions in capturing human travelers' complex learning and adjustment behavior. Although the memory-only LLM baseline showed some predictive capability, likely by inferring patterns from the choice history, the superior performance of our approach validates the strategy of explicitly learning a persona. This confirms that a data-driven, calibrated persona provides a more accurate and effective foundation for simulation, a conclusion supported by related findings in static choice contexts in transportation \citep{liu2024can,liu2025aligning}.

To gain more information on the specific performance gain of our approach, we compare the F1 score of our proposed approach with the baseline approaches in \Cref{tab:agent_f1_comp}. In the table, the number in parentheses is the ranking of each method on the corresponding agent.

\begin{table}[h!]
	\caption{Simulation F1-score for each agent}\label{tab:agent_f1_comp}
	\begin{center}
		\begin{tabular}{l c c c c c }
        \hline
        OD group & ID & 
        Ours & Base-LLM & Recursive-LLM & Bounded-LLM\\
        \hline
        \multirow{9}{*}{Group 1}& 1 & 0.514(3) & 0.450(4) & 0.657(\textbf{1}) & 0.630(2)\\ 
        & 2 & 0.527(2) & 0.453(4) & 0.533(\textbf{1}) & 0.502(3) \\
        & 3 & 0.557(2) & 0.620(\textbf{1}) & 0.551(3) & 0.450(4)\\
        & 4 & 0.825(\textbf{1}) & 0.630(2) & 0.112(3) & 0.018(4)\\
        & 5 & 0.685(\textbf{1}) & 0.506(2) & 0.479(3) & 0.413(4)\\
        & 6 & 0.581(2) & 0.652(\textbf{1}) & 0.291(4) & 0.306(3)\\
        & 7 & 0.695(3) & 0.631(4) & 0.731(\textbf{2}) & 0.732(\textbf{1})\\
        & 8 & 0.599(\textbf{1}) & 0.354(2) & 0.342(3) & 0.334(4)\\
        & 9 & 0.582(2) & 0.398(4) & 0.649(3) & 0.678(\textbf{1})\\ 
        \hdashline
        \multirow{6}{*}{Group 2} & 10 & 0.889(\textbf{1}) & 0.871(2) & 0.010(4) & 0.013(3)\\
        & 11 & 0.981(\textbf{1}) & 0.830(2) & 0.025(3) & 0.025(4)\\
        & 12 & 0.544(2) & 0.569(\textbf{1}) & 0.302(4) & 0.344(3)\\
        & 13 & 0.520(\textbf{1}) & 0.426(4) & 0.428(3) & 0.471(2)\\
        & 14 & 0.463(\textbf{1}) & 0.292(4) & 0.396(3) & 0.443(2)\\
        & 15 & 0.532(2) & 0.584(\textbf{1}) & 0.362(4) & 0.389(3)\\
        \hline
    \multicolumn{2}{l}{Average rank} & \textbf{1.67} & 2.60 & 3.00 & 2.73\\
\multicolumn{2}{l}{Average gap to \#1} & \textbf{0.046} & 0.111 & 0.267 & 0.264\\
\hline
		\end{tabular}
	\end{center}
\end{table}\par

An analysis of individual agent performance, presented in \Cref{tab:agent_f1_comp}, further underscores the robustness of our approach. Our method consistently achieves high rankings for all agents and is never the worst-performing model. In contrast, the baselines show significant performance fluctuations across the agent pool. Moreover, in cases where our model is not the top performer, it remains highly competitive with a small performance gap. This stability is reflected in a mean rank of 1.67, substantially better by almost one rank than the next-best baseline (2.53), and translates to a 58.6\% reduction in the average performance gap to the top-ranked model (from 0.111 to 0.046). A granular analysis of per-agent performance reveals how different modeling approaches handle specific behavioral patterns. For participants with stable but mixed strategies, the theory-driven models perform well. However, their rigid assumptions fail when applied to agents who exhibit strong route biases (e.g., 4, 10, 11). The Base-LLM shows the opposite weakness: while sometimes competitive, it often fails to accurately model agents with balanced choice patterns (e.g., 1, 2, 9, 13). Our proposed approach distinguishes itself by achieving high accuracy across both of these behavioral types. The most significant differentiator, however, is in capturing dynamic learning. Our method consistently outperforms baselines for agents who exhibited non-stationary behavior by substantially changing their strategies (2, 5, 8, 13, 14, 15). This provides strong evidence that our framework successfully learns the evolution of an agent's strategy, a critical capability for realistic simulation.

The learned personas of agents reflect their decision-making tendencies, and the differences in each agent can also be reflected in the learned personas. Some learned personas of representative agents are shown in \Cref{tab:agent_personas_long}:

\begin{longtable}{p{0.02\textwidth} p{0.98\textwidth}}
	
	\caption{Example of learned personas of agents}
	\label{tab:agent_personas_long} \\
	
	\hline
	\textbf{ID} & \textbf{Learned persona} \\
	\hline
	\endfirsthead
	
	\multicolumn{2}{c}%
	{{\bfseries\tablename\ \thetable{} -- continued from previous page}} \\
	\hline
	\textbf{ID} & \textbf{Learned persona} \\
	\hline
	\endhead
	
	\hline
	\multicolumn{2}{r}{{Continued on next page}} \\
	\endfoot
	
	\hline
	\endlastfoot
	
	6 & \textbf{Short-Term Processing Rule}:  
Prioritize recent experiences by evaluating the travel times from the last 5 trips. Favor routes that consistently deliver shorter travel times, but avoid routes that have shown extreme or unpredictable delays in the recent past (e.g., significantly longer durations compared to the average).

\textbf{Long-Term Strategy Rule}: 
Maintain a default bias toward routes that have historically provided the most consistent and reasonable travel times. Be cautious of routes with a history of frequent extreme delays, even if they occasionally offer faster times.

\textbf{Decision Heuristic}:  
Combine short-term and long-term rules by weighting recent trip performance more heavily when making immediate decisions, while adjusting for historical patterns of reliability. Choose the route that balances shorter recent travel times with long-term consistency, avoiding routes with high variability or risk of significant delays.\\
	3 & \textbf{Short-Term Processing Rule:} 
	Monitor the last 3-5 trips on the current route. Switch to the alternative route if delays exceed 25 minutes in two consecutive trips \textit{or} variability exceeds 15 minutes across three trips, unless recovery trends are evident (e.g., two consecutive trips with delays below 20 minutes or consistent improvement over three trips). Treat delays exceeding 50 minutes as critical signals of instability, but avoid immediate reactions unless instability persists across multiple trips. Favor routes showing travel times consistently under 20 minutes, but temper this preference by considering one-off shorter travel times (e.g., under 20 minutes) as indicators only if aligned with broader stability across recent trips.
	
	\textbf{Long-Term Strategy Rule:} 
	Maintain preference for routes with historically faster average travel times and lower variability, while penalizing recurring extreme delays ($>$50 minutes) observed in at least three trips within a recent 10-trip window. Reward recovery trends when recent performance demonstrates stabilization, and prioritize routes showing consistent travel times and low variability over longer periods. Avoid reverting to historically favorable routes unless recent data confirms reliability and consistency. Default biases toward expressways or other historically faster routes should be moderated by higher local distribution trends and consistent performance, ensuring alignment with broader reliability indicators.
	
	\textbf{Decision Heuristic:} 
	Combine short-term and long-term rules with dynamic weighting: assign 60\% weight to long-term trends and 40\% weight to short-term data, increasing short-term weight only when recovery trends are clear or instability persists (e.g., sustained delays or variability over three or more trips). Ensure transient disruptions or isolated events do not disproportionately influence decisions, while factoring in significantly shorter travel times (e.g., under 20 minutes) as secondary indicators only if consistent with broader trends. Integrate traffic distribution trends, favoring routes with higher commuter traffic only when aligned with stable and consistent performance.

	\\ \midrule
	
	7 & \textbf{Short-Term Processing Rule}:  
Evaluate the average travel time across the last 5 trips, prioritizing routes with consistently lower averages. Avoid overreacting to isolated deviations unless they persist for at least 3 consecutive trips. Incorporate the travel time of the most recent trip as supplementary information by assigning a weight: negative if greater than 20 minutes, positive if less than 15 minutes, and neutral if between 15 and 20 minutes. Use these weights to refine the broader pattern evaluation.

\textbf{Long-Term Strategy Rule}: 
Maintain a default bias toward the expressway, as historical data confirms its faster travel times under favorable conditions. Define "consistent poor performance" as the expressway exceeding a specific threshold (e.g., 20 minutes) for 5 consecutive trips or averaging over 30 minutes in recent consecutive trips. Switch to local1 only if it demonstrates sustained improvement over 5 consecutive trips with an average travel time significantly lower than the expressway. Revert to the expressway once conditions stabilize or variability decreases.

\textbf{Decision Heuristic}:  
Balance short-term trends and long-term data by prioritizing the expressway when trends are mixed or inconclusive. Switch to local1 only if both short-term averages and long-term data clearly indicate persistent poor performance on the expressway and sustained improvement on local1. If the short-term weight for the most recent trip is negative and aligns with broader data showing consecutive poor expressway performance, prefer local1 for the next trip. Recent data should only outweigh long-term trends if it shows sustained, extreme deviation over at least 5 consecutive trips, ensuring optimal travel time while minimizing risk.
	\\ \midrule
	
	11 & \textbf{Short-Term Processing Rule}:
Prioritize the most recent trips by evaluating the travel time of the last few choices. If my most recent route consistently results in travel times close to or better than my historical average (around 15.9 minutes for 'local2'), I should favor repeating that choice. If my latest choice deviates significantly or results in worse outcomes, reconsider the alternative.

\textbf{Long-Term Strategy Rule}:
Avoid high-risk options based on past extreme negative experiences. The 'expressway' route is inherently biased as unreliable due to historical outliers (e.g., 72.8 minutes), even if it occasionally offers competitive travel times. My default belief is that 'local2' is the safer, more consistent option, with predictable and moderate travel times.

\textbf{Decision Heuristic}:
Combine short-term validation with long-term bias to make a final choice. If the short-term travel times for 'local2' remain consistent and below historical risk thresholds (e.g., avoiding spikes like 17.2 minutes or worse), I should choose 'local2.' Only consider 'expressway' if recent data strongly suggests a significant and reliable improvement in travel time distribution without extreme risks.
	\\ 
\end{longtable}

The learned personas of agents reflect their decision-making tendencies, and the differences in each agent can also be reflected in the learned personas. \Cref{tab:agent_personas_long} shows the learned personas of four representative agents: agents 3 and 6, who maintain a relatively balanced distribution between the two routes, and agents 7 and 11, whose choices are heavily inclined towards one option. The learned personas effectively capture these distinct tendencies by articulating the agents' decision-making priorities. For agents 3 and 6, their personas reflect a careful consideration of both short- and long-term memory to evaluate route performance and reliability. Their decision-making prioritizes a rational evaluation of travel time and is not biased towards one particular route. These underlying mechanisms drive them to react to changing conditions and switch routes strategically to minimize their travel times. In contrast, the learned personas of agents 7 and 11 reveal a strong cognitive bias to default to one route. Agent 7's persona states it will "Switch to local1 only if it demonstrates sustained improvement over 5 consecutive trips," while agent 11's contains the belief that `local2' is the safer, more consistent option." While both agents will deviate from their default choice when evidence is overwhelmingly clear, their personas enforce a high threshold for switching and a strong inclination to revert to their preferred option. Overall, the analysis confirms that our framework generates good and interpretable models, successfully capturing a spectrum of learning and decision-making styles in travelers' day-to-day behavior.

Also, as evidenced in \Cref{tab:agent_f1_comp}, our dual-agent approach can learn the agent's gradual update of their personas as they interact with the system and obtain more experiences, therefore helping out-of-sample generalization. This process is also evident in the agent's persona change in the learning process, as we show by using the example of agent 8 in \Cref{tab:agent_personas_progress}.

\begin{longtable}{p{0.05\textwidth} p{0.95\textwidth}}
	
	\caption{The starting and ending persona of agent 8}
	\label{tab:agent_personas_progress} \\
	
	\hline
	\textbf{Time} & \textbf{Learned persona} \\
	\hline
	\endfirsthead
	
	\multicolumn{2}{c}%
	{{\bfseries\tablename\ \thetable{} -- continued from previous page}} \\
	\hline
	\textbf{ID} & \textbf{Learned persona} \\
	\hline
	\endhead
	
	\hline
	\multicolumn{2}{r}{{Continued on next page}} \\
	\endfoot
	
	\hline
	\endlastfoot
	
	
	Start of training & \textbf{Short-Term Processing Rule}:
Prioritize recent experiences by heavily weighing the travel times from the last 5 trips. If a route consistently performs better in terms of shorter travel time within this window, favor that route. However, penalize routes that result in significantly higher-than-average travel times (e.g., exceeding 50 minutes), even if they performed well during other recent trips.

\textbf{Long-Term Strategy Rule}:
Default to believing the expressway is generally faster and more reliable, unless historical data shows repeated extreme delays (e.g., over 70 minutes). Assume that lower travel time is more likely on the expressway, especially when the distribution favors expressway usage over local routes.

\textbf{Decision Heuristic}:
Combine short-term and long-term rules by first evaluating the recent performance (short-term rule) and then adjusting the decision based on long-term bias. Choose the expressway if recent travel times are comparable or better than local routes, but switch to local routes if the expressway has shown consistently poor performance or if recent data shows a significant risk of long delays. Prioritize avoiding extreme delays over adherence to historical bias.
	\\ \midrule 
	
	End of training & \textbf{Short-Term Processing Rule}:
Evaluate the last 5 trips with a focus on identifying sustained patterns rather than isolated events. Prioritize routes with short-term stability and consistently moderate travel times, rewarding reliable performance even if occasional variability is present, provided extreme delays are infrequent and do not form clear worsening trends. Avoid penalizing routes for isolated negative events unless extreme delays occur repeatedly over at least 3 consecutive trips. Ensure negative patterns are assessed with proportional penalties, emphasizing overall consistency and avoiding overreaction to anomalies.

\textbf{Long-Term Strategy Rule}:
Favor routes with faster historical averages and sustained reliability, maintaining a strong bias toward options that consistently demonstrate long-term superiority. Penalize extreme delays ($>$50 minutes) only if they occur repeatedly and disrupt long-term reliability. Balance flexibility to shift toward alternatives when sustained improvements emerge with maintaining preference for historically faster options. Minimize short-term disruptions from isolated events unless they clearly impact long-term reliability or show sustained worsening patterns.

\textbf{Decision Heuristic}:
Combine short-term trends and long-term reliability, ensuring decisions emphasize sustained advantages across both dimensions. Prioritize routes that demonstrate consistent reliability and faster historical averages while ensuring recent performance reflects broader patterns rather than isolated anomalies. Slightly favor long-term superiority when historical trends clearly outweigh short-term disruptions. Apply penalties proportionally, balancing rewards for sustained reliability and moderate travel times across both short-term and long-term considerations without overreacting to isolated negative events.\\
    \hline
\end{longtable}

The evolution of agent 8's persona in \Cref{tab:agent_personas_progress} provides a clear longitudinal example of the calibration process. The agent begins with a naive and highly reactive strategy, heavily weighing the most recent trips and over-penalizing singular extreme delays, all while relying on a simple default bias for the expressway. Through the training process, this persona matures into a significantly more sophisticated and robust policy. The final persona explicitly demonstrates that the agent has learned to differentiate signal from noise: it now prioritizes "sustained patterns" and "long-term reliability" while actively "avoiding overreaction to anomalies" and "isolated negative events". Consequently, the agent's behavior evolves from heavily favoring the expressway in the first 20 days to having balanced choices between routes in the last 20 days. This progression of persona provides an example of the training process in action, validating our adaptive learning mechanism.

\begin{table}[h!]
	\caption{Cosine similarity of simulated and human behavior vectors for each agent}\label{tab:agent_cosine_comp}
	\begin{center}
		\begin{tabular}{l c c c c c }
        \hline
        OD group & ID & 
        Ours & Base-LLM & Recursive-LLM & Bounded-LLM\\
        \hline
        \multirow{9}{*}{Group 1}& 1 & 0.880(4) & 0.921(3) & 0.942(2) & 0.968(\textbf{1})\\ 
        & 2 & 0.948(4) & 0.995(\textbf{1}) & 0.989(2) & 0.986(3) \\
        & 3 & 0.980(3) & 0.938(4) & 0.998(\textbf{1}) & 0.983(2)\\
        & 4 & 1.000(\textbf{1}) & 0.799(4) & 0.993(3) & 0.996(2)\\
        & 5 & 0.990(\textbf{1}) & 0.887(2) & 0.777(4) & 0.817(3)\\
        & 6 & 1.000(\textbf{1}) & 1.000(\textbf{1}) & 0.998(4) & 1.000(\textbf{1})\\
        & 7 & 0.851(4) & 0.999(\textbf{1}) & 0.929(2) & 0.916(3)\\
        & 8 & 0.999(2) & 0.940(4) & 1.000(\textbf{1}) & 0.999(2)\\
        & 9 & 0.999(1) & 0.987(4) & 0.999(2) & 0.999(3)\\ 
        \hdashline
        \multirow{6}{*}{Group 2} & 10 & 1.000(\textbf{1}) & 0.977(4) & 1.000(\textbf{1}) & 1.000(\textbf{1})\\
        & 11 & 1.000(\textbf{1}) & 0.949(4) & 1.000(\textbf{1}) & 0.999(3)\\
        & 12 & 0.977(2) & 0.993(\textbf{1}) & 0.943(3) & 0.927(4)\\
        & 13 & 0.998(\textbf{1}) & 0.944(4) & 0.984(3) & 0.993(2)\\
        & 14 & 0.997(2)& 0.999(\textbf{1}) & 0.958(4) & 0.972(3)\\
        & 15 & 0.988(2) & 0.848(4) & 0.930(3) & 0.990(\textbf{1})\\
        \hline
    \multicolumn{2}{l}{Average rank} & \textbf{2.00} & 2.80 & 2.33 & 2.13\\
\multicolumn{2}{l}{Average value} & \textbf{0.974} & 0.945 & 0.963 & 0.970\\
\hline
		\end{tabular}
	\end{center}
\end{table}\par

We first present a quantitative evaluation of behavioral fidelity in \Cref{tab:agent_cosine_comp}. This table reports the cosine similarity between the simulated behavior vector $(C^{-}, S^{-})$ and the human ground truth for all 15 agents. The number in parentheses indicates the performance rank of each method for that specific agent, from (1) for best (highest similarity) to (4) for worst.

The summary statistics at the bottom confirm the superior performance of our framework. Our method achieves the best average rank (2.00) and the highest average value (0.974), indicating it most consistently and accurately replicates the true human behavioral vectors. The Bounded-LLM is the closest competitor (average rank 2.13, average value 0.970), while the Base-LLM performs the poorest on both metrics. Notably, our method achieves a perfect similarity score of 1.000 for four different agents (agents 4, 6, 10, and 11), demonstrating robust alignment in numerous cases. While this quantitative data confirms that our method is the most accurate, the following four-corner analysis will visually explain why by decomposing these vectors.

\begin{figure}[h]
    \centering
    \includegraphics[width=0.95\linewidth]{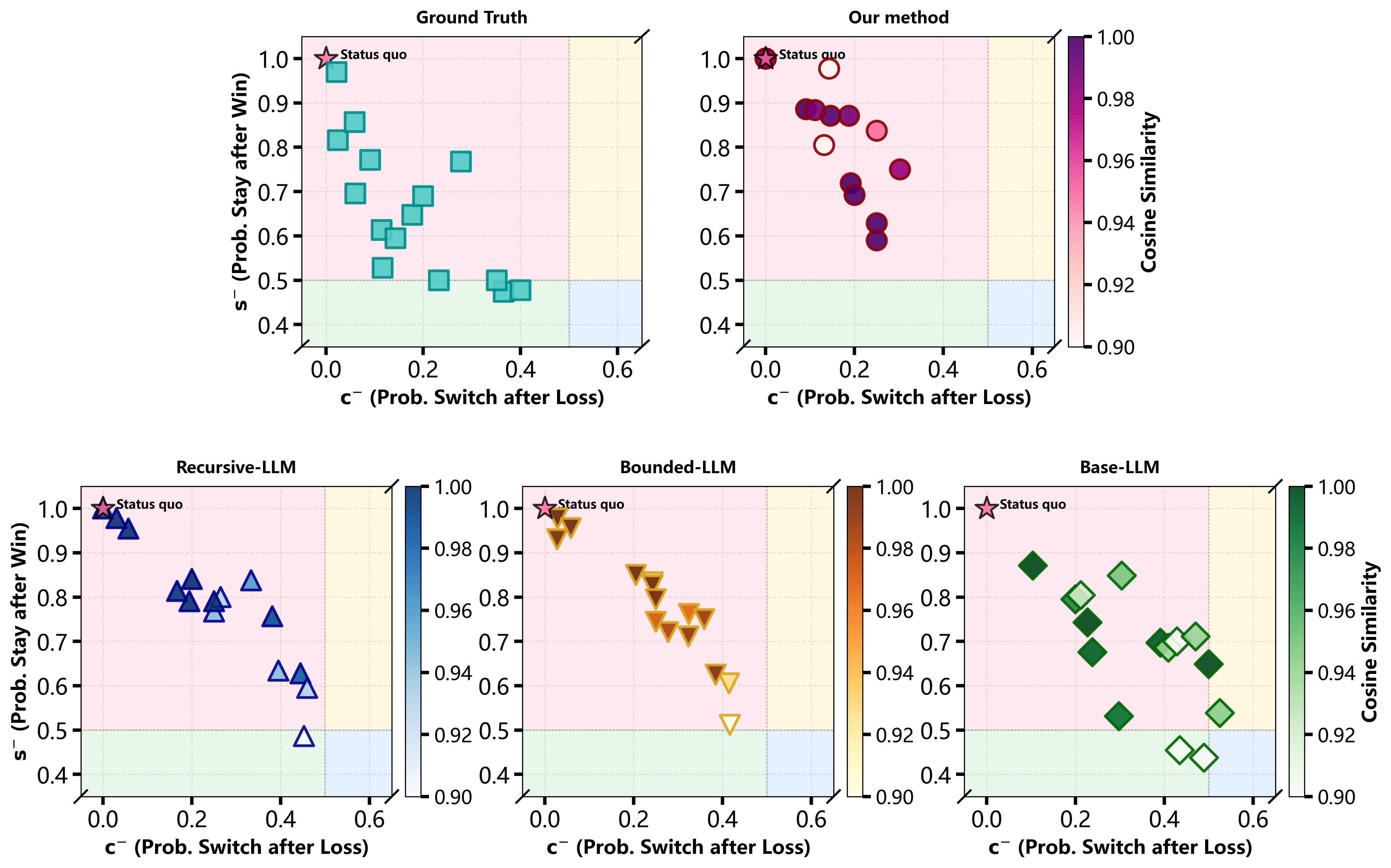}
    \caption{
        Global behavior vector distribution (Day 20–39) across methods. Marker color intensity reflects cosine similarity to the corresponding human traveler.
    }
    \label{fig:behavior_vectors_global}
\end{figure}

\Cref{fig:behavior_vectors_global} visualizes the global distribution of these vectors (days 20–39) across methods for the Ground Truth against the four LLM methods. The plots reveal a clear distinction. For our framework (Purple Circles), the markers have mostly dark colors (indicating values $\approx 1.0$), showing high cosine similarity with humans, which confirms this strong quantitative alignment. This suggests that our dual-agent architecture effectively models the human separation between learning and behavioral adjustment. In contrast, the baseline methods fail. The Base-LLM (green diamonds) agents are erratically distributed and exhibit low cosine similarity. While the Recursive-LLM (blue triangles) and Bounded-LLM (orange inverted triangles) agents correctly learn the high $S^{-}$ component, they fundamentally misunderstand inertia. Both incorrectly exhibit much higher $C^{-}$ values. In conclusion, \Cref{fig:behavior_vectors_global} demonstrates that baseline LLM approaches tend towards an overly reactive "lose-shift" behavior. Our framework is the only one to uniquely model human behavioral inertia, achieving superior alignment with the human travelers.

\subsection{Comparison of system dynamics}

\begin{figure}[htbp]
\centering
    \subfloat[][Expressway]{\includegraphics[width=0.85\linewidth]{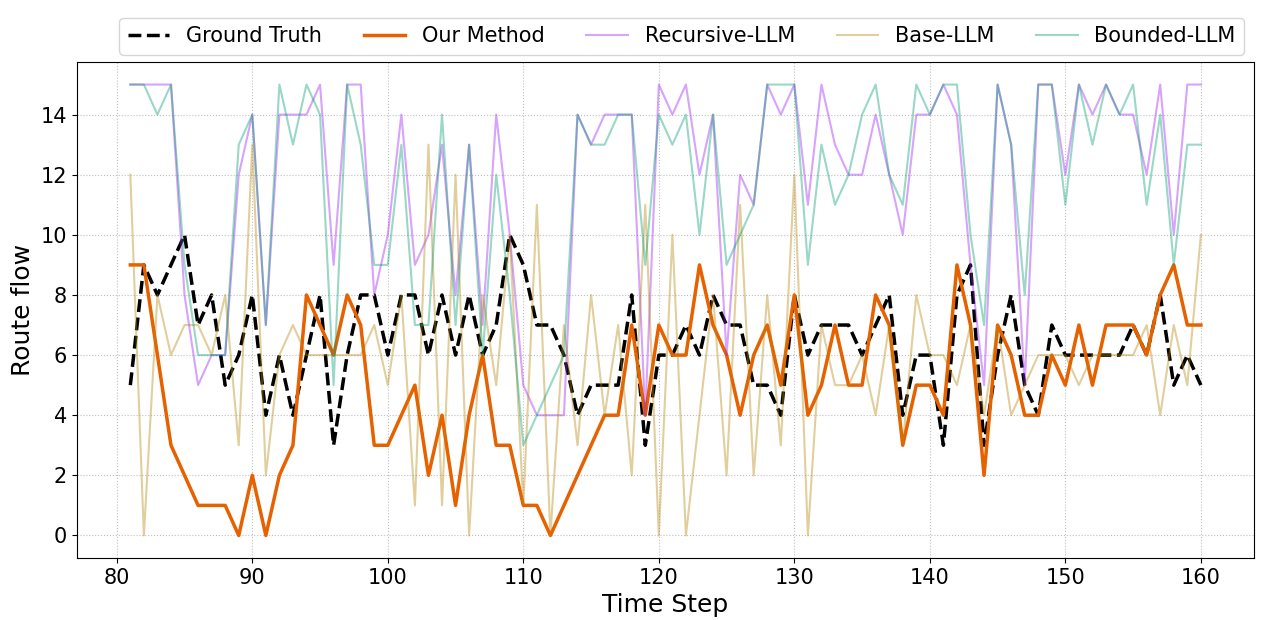}}
\newline
	\subfloat[][Arterial ``local 1"]{\includegraphics[width=0.85\linewidth]{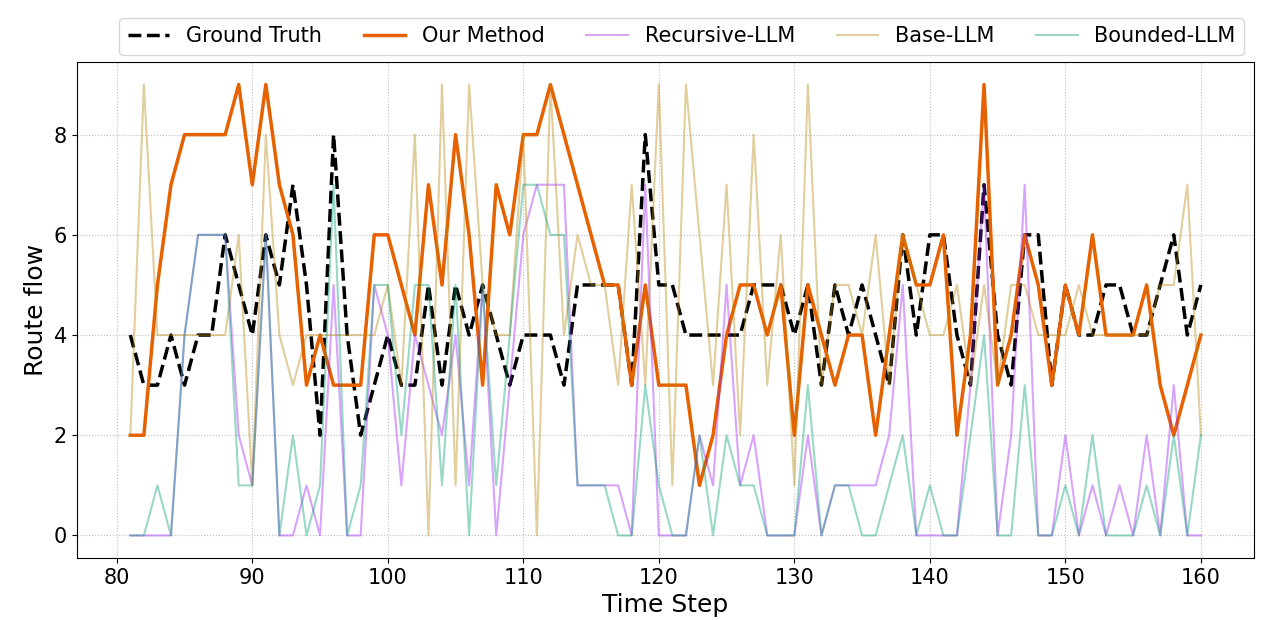}}
	\newline
    \subfloat[][Arterial ``local 2"]{\includegraphics[width=0.85\linewidth]{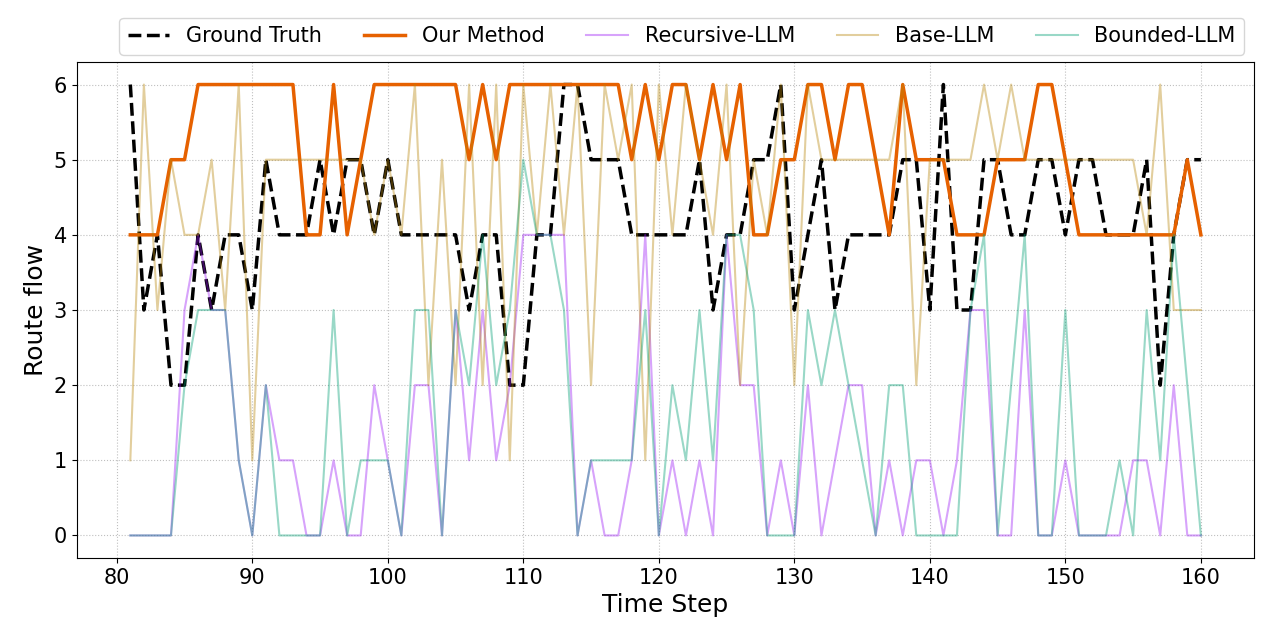}}
	\caption{Simulated route flows under controlled system dynamics}
	\label{fig:simulated_flow_controlled}
\end{figure}

In this section, we compare the aggregate simulation performance of our methods and established approaches over the simulated route flow. First, we control the system dynamics by providing all models with the ground truth system state from the previous days, and simulate the route flow progression for the last 80 days. This approach is designed to test the overall route choice tendencies of the models while isolating their aggregate behavior from the confounding effects of compounding simulation errors. The simulation results are presented in \Cref{fig:simulated_flow_controlled}, while evaluations of the simulation results are presented in \Cref{tab:control_route_flow_comp}.

\begin{table}[h!]
	\caption{Model comparison for aggregate route flow simulation under controlled system dynamics}\label{tab:control_route_flow_comp}
	\begin{center}
		\begin{tabular}{l c c }
        \hline
        & MAPE & MSE \\
        \hline
        Base-LLM & 42.3\% & 7.49\\
        Recursive-LLM & 81.3\% & 22.36\\
        Bounded-LLM & 78.9\% & 21.24\\
        Ours &  40.0\% & 5.94  \\
\hline
		\end{tabular}
	\end{center}
\end{table}\par

The simulation results, presented in \Cref{fig:simulated_flow_controlled}, demonstrate our model's ability to accurately reproduce system-level route flows. Qualitatively, our model's simulated flow not only converges with the ground truth, particularly in the latter half of the simulation (days 120+), but also mirrors its characteristic smoothness. This contrasts sharply with the Base-LLM, which produces a highly volatile flow, and the other two baselines, which fail to capture the correct trend. This suggests their simpler decision mechanisms or recursive structures are not well-suited for leveraging accurate state information, potentially leading to over-reactions and chaotic system behavior. These qualitative findings are supported by quantitative metrics in \Cref{tab:control_route_flow_comp}, where our method reduces the MAPE by 5.4\% (from 42.3\% to 40.0\%) and the MSE by 20.7\% (from 7.49 to 5.94) relative to the best-performing baseline. Therefore, our model excels not only in its quantitative accuracy but also in its qualitative ability to replicate the realistic, low-volatility dynamics of the ground-truth route flows.

\begin{figure}[htbp]
\centering
    \subfloat[][Expressway]{\includegraphics[width=0.85\linewidth]{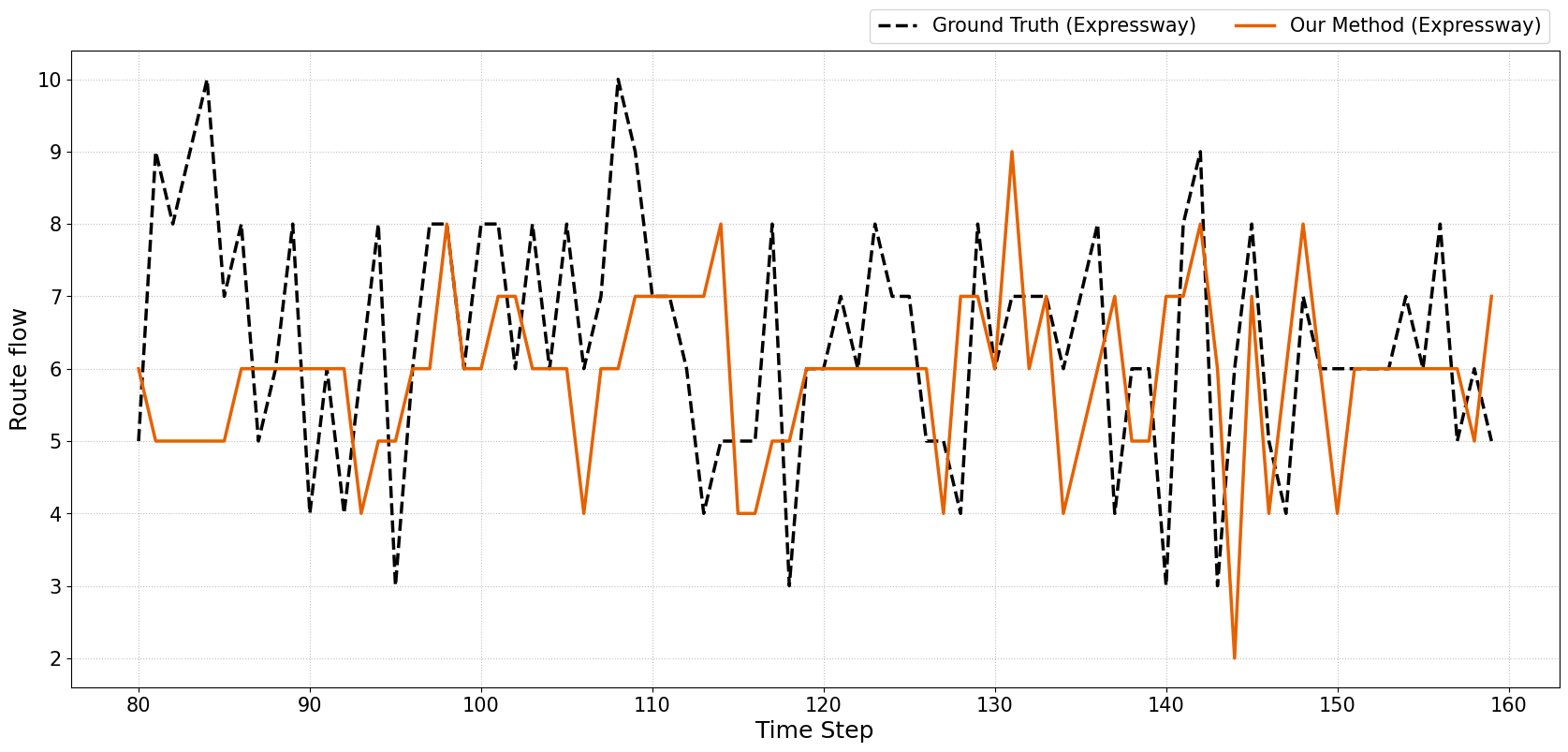}}
\newline
	\subfloat[][Arterial ``local 1"]{\includegraphics[width=0.85\linewidth]{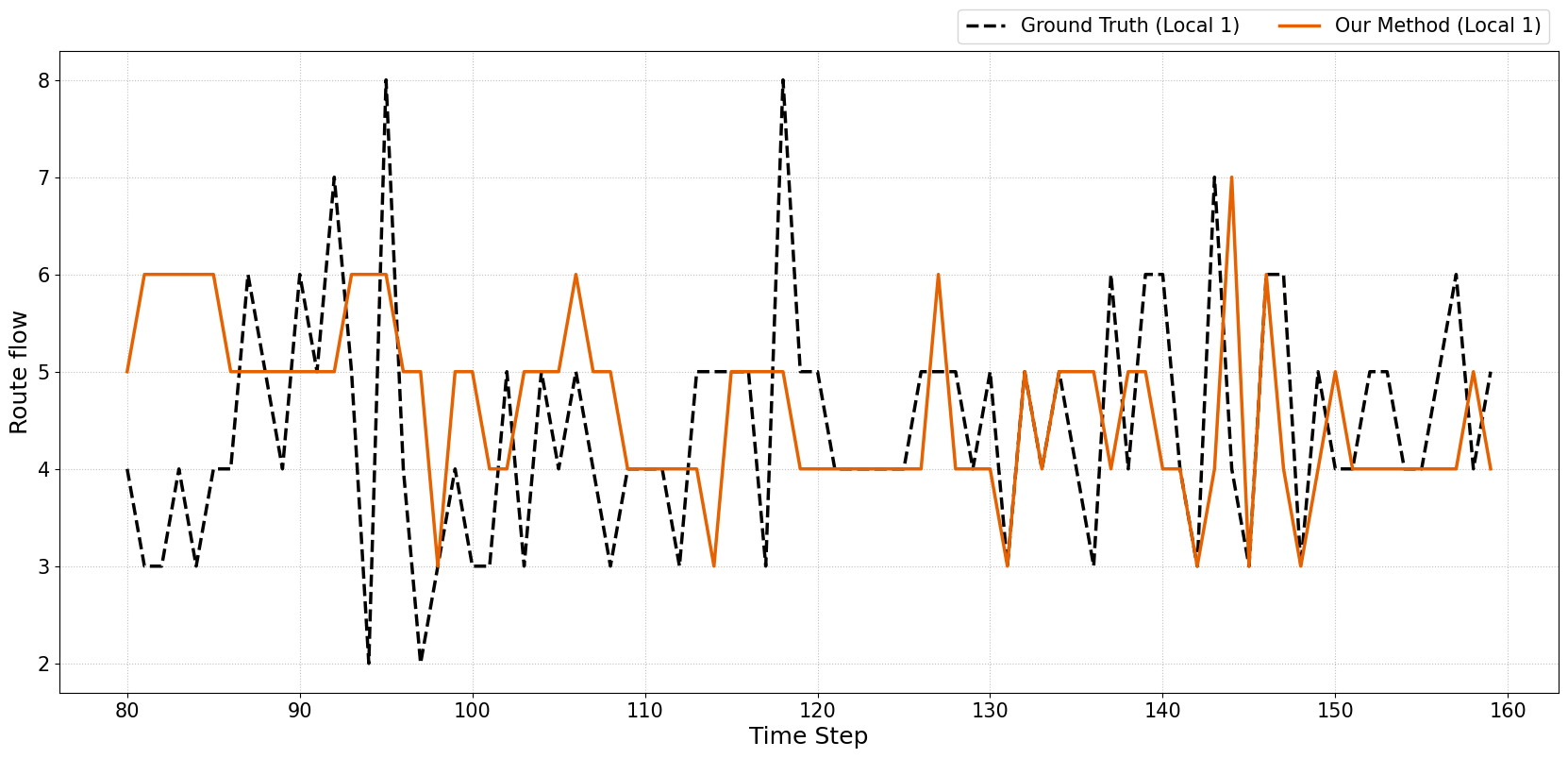}}
	\newline
    \subfloat[][Arterial ``local 2"]{\includegraphics[width=0.85\linewidth]{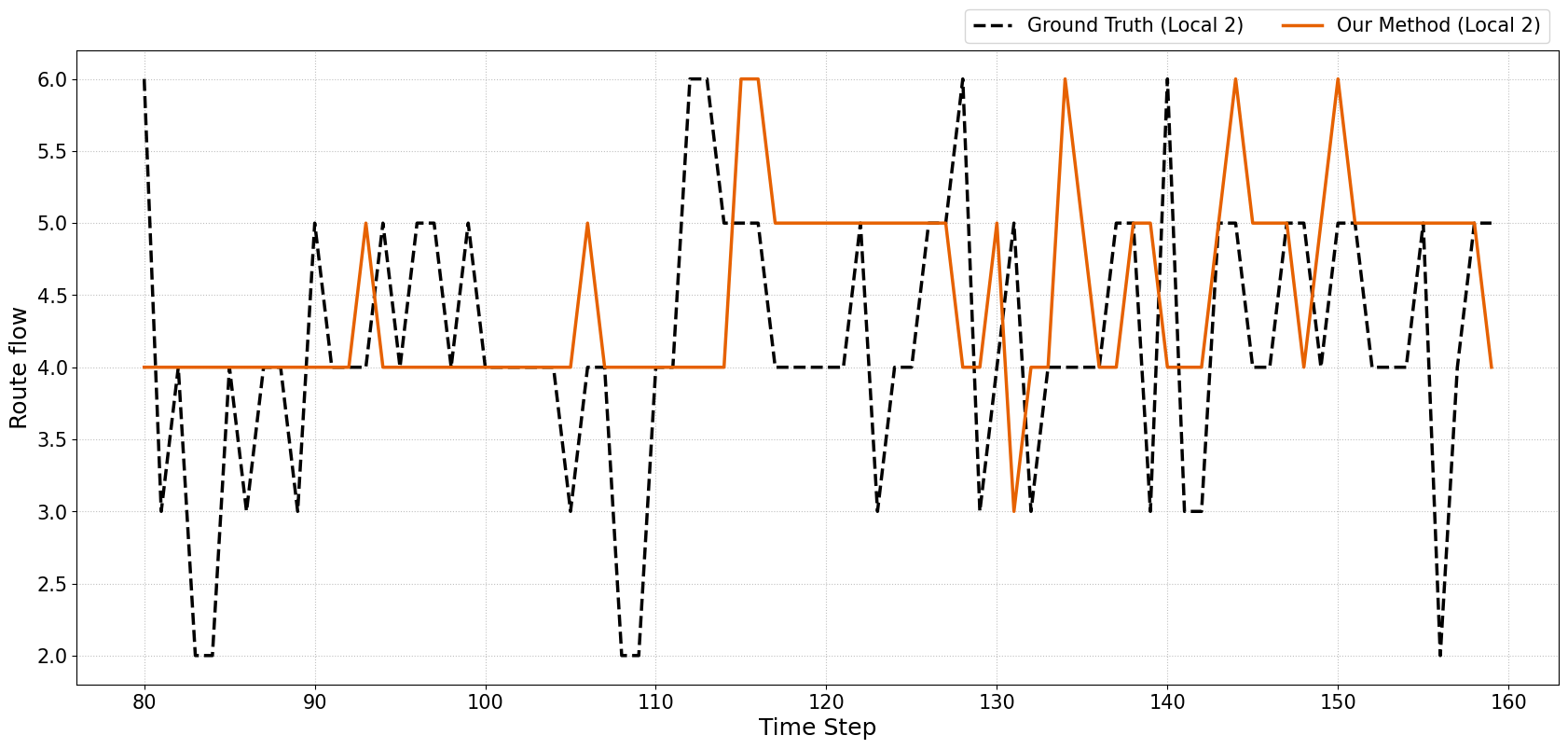}}
	\caption{Simulated route flows under uncontrolled system dynamics}
	\label{fig:simulated_flow_not_controlled}
\end{figure}

Next, we evaluate the performance of our approach by conducting an experiment that removes the controls in \Cref{fig:simulated_flow_controlled} to resemble standard counterfactual simulations. In this setting, agents learn and adapt based solely on the travel times they experience, processed through their memory mechanism, without access to the ground truth flows from the previous step. This represents a more realistic scenario characterized by imperfect information and experiential learning.

\Cref{fig:simulated_flow_not_controlled} provides an isolated comparison between the route flows generated by our model under uncontrolled system dynamics (orange line) and the ground truth (dashed black line). The simulation successfully captures the qualitative trends and persistent fluctuations of the ground truth, demonstrating effective adaptive learning. However, the simulated flows exhibit occasional lags or deviations. This is likely because the agents are learning from inherently noisy system dynamics and face delays in reacting to changes, as they rely solely on experience rather than perfect state information. Despite this reduced moment-to-moment accuracy, the model's ability to reproduce these key dynamic characteristics validates its capacity to approximate system behavior using only experiential feedback.


\section{Conclusion} \label{sec:conclusion}

In this paper, we introduce a novel dual-agent framework to automatically align LLM agents with human travelers for simulating the dynamic learning and adjustment of humans in transportation systems. Our approach features a set of LLM traveler agents, which use a memory system and a learnable persona to simulate the learning and subsequent decision-making of human travelers, and a supervisory LLM calibration agent which performs an automated, online optimization of the traveler's persona using a textual ``pseudo-gradient" descent method. Using an online automatic simulation and calibration procedure, we transform the complex task of behavioral alignment into an iterative, rolling-window optimization problem that is jointly solved by the simulations of LLM traveler agents and the learning updates of the LLM calibration agent. Our empirical evaluation on a real-world day-to-day route choice dataset demonstrates that our framework significantly outperforms existing baselines in both individual learning accuracy and aggregate traffic flow simulation, the former by the F1-score by 16.7\%, and the latter by reducing the route flow MSE by 20.7\%. Furthermore, our analysis of the model performance on individual travelers reveals the method's ability to align not just with behavioral outcomes, but with the underlying decision-making tendencies and their evolution. We observe that our model significantly outperforms baseline models when the traveler exhibits significant changes in their decision-making strategies with time, and the assessment of the learned personas reveals that our approach can explicitly capture such changes in persona. Furthermore, using a decision-making tendency evaluation in \cite{wang2025comparing}, our model also aligns the best with human ground-truth, further strengthening the alignment. Computationally, we save the number of tokens needed for the training and evaluation process by summarizing all learned behavior into a compact persona and a fixed-length memory for the traveler agents, and use a small rolling window for the calibration agents. This strategy reduces computational overhead, making the alignment process more cost-effective.

Future research can build on this study by focusing on several promising directions. First, enhancing the computational efficiency of the online calibration process is a key area for development. This could involve exploring more token-efficient methods for aggregating pseudo-gradients, such as developing single-pass ``batch" updates, or refining the candidate generation strategy to reduce the number of LLM queries required per update, thereby lowering computational costs and latency. Second, future work could explore the joint learning of both the persona and the memory retrieval mechanisms. While we fixed these retrieval heuristics for identifiability and learning stability, allowing the agent to also learn its own information-weighting and memory utilization strategies could create agents with even more adaptive and personalized decision-making styles.

\section*{Acknowledgement}

The work described in this paper was partly supported by research grants from National Science Foundation (CMMI-2233057 and 2240981).

\newpage
\bibliographystyle{apalike}


\end{document}

%% file: Premable.tex
\documentclass[11pt,english]{article}
\usepackage[sc]{mathpazo}
\usepackage{geometry}
\usepackage{color}
\usepackage{array}
\usepackage{bbding}
\usepackage{multirow}
\usepackage[fleqn]{amsmath}
\usepackage{amssymb}
\usepackage{amsthm}
\usepackage{graphicx}
\usepackage{natbib}
\usepackage{lscape}
\usepackage[hidelinks]{hyperref}
\usepackage{comment}
\usepackage{bm}
\usepackage[title]{appendix}
\usepackage{longtable}
\usepackage{mathtools}
\usepackage{chngcntr}
\usepackage{authblk}
\usepackage{siunitx}
\usepackage{babel}
\usepackage{dsfont}
\usepackage[most]{tcolorbox}
\usepackage[short]{optidef}
\usepackage{algorithm}
\usepackage{algpseudocode}
\usepackage{arydshln}

\usepackage{enumitem}

\usepackage[most]{tcolorbox} 
\usepackage[hidelinks]{hyperref}  

\makeatletter
\allowdisplaybreaks
\usepackage[mathlines, pagewise]{lineno}
\usepackage{etoolbox} 

\newcommand*\linenomathpatchAMS[1]{%
	\expandafter\pretocmd\csname #1\endcsname {\linenomathAMS}{}{}%
	\expandafter\pretocmd\csname #1*\endcsname{\linenomathAMS}{}{}%
	\expandafter\apptocmd\csname end#1\endcsname {\endlinenomath}{}{}%
	\expandafter\apptocmd\csname end#1*\endcsname{\endlinenomath}{}{}%
}

\expandafter\ifx\linenomath\linenomathWithnumbers
\let\linenomathAMS\linenomathWithnumbers
\patchcmd\linenomathAMS{\advance\postdisplaypenalty\linenopenalty}{}{}{}
\else
\let\linenomathAMS\linenomathNonumbers
\fi

\linenomathpatchAMS{gather}
\linenomathpatchAMS{multline}
\linenomathpatchAMS{align}
\linenomathpatchAMS{alignat}
\linenomathpatchAMS{flalign}

\@ifundefined{showcaptionsetup}{}{%
 \PassOptionsToPackage{caption=false}{subfig}}
\usepackage{subfig}
\makeatother

\usepackage{babel}

\newcolumntype{L}[1]{>{\RaggedRight\arraybackslash}p{#1}}

\tcbset{
  promptstyle/.style={
    breakable,
    colback=white,
    colframe=black!20,
    fonttitle=\bfseries,
    list inside=lop,
    list entry={Prompt \thetcbcounter: #1}, 
  }
}

\newtcolorbox[auto counter]{promptbox}[2][Omm]{
  promptstyle, 
  title={#2},  
  #1           
}

%% file: main.bbl
\begin{thebibliography}{}

\bibitem[Achiam et~al., 2023]{achiam2023gpt}
Achiam, J., Adler, S., Agarwal, S., Ahmad, L., Akkaya, I., Aleman, F.~L., Almeida, D., Altenschmidt, J., Altman, S., Anadkat, S., et~al. (2023).
\newblock Gpt-4 technical report.
\newblock {\em arXiv preprint arXiv:2303.08774}.

\bibitem[Adornetto et~al., 2025]{adornetto2025generative}
Adornetto, C., Mora, A., Hu, K., Garcia, L.~I., Atchade-Adelomou, P., Greco, G., Pastor, L. A.~A., and Larson, K. (2025).
\newblock Generative agents in agent-based modeling: Overview, validation, and emerging challenges.
\newblock {\em IEEE Transactions on Artificial Intelligence}.

\bibitem[Alsaleh and Farooq, 2025]{alsaleh2025towards}
Alsaleh, T. and Farooq, B. (2025).
\newblock Towards locally deployable fine-tuned causal large language models for mode choice behaviour.
\newblock {\em arXiv preprint arXiv:2507.21432}.

\bibitem[Avineri and Prashker, 2005]{avineri2005sensitivity}
Avineri, E. and Prashker, J.~N. (2005).
\newblock Sensitivity to travel time variability: travelers’ learning perspective.
\newblock {\em Transportation Research Part C: Emerging Technologies}, 13(2):157--183.

\bibitem[Baddeley and Hitch, 1993]{baddeley1993recency}
Baddeley, A.~D. and Hitch, G. (1993).
\newblock The recency effect: Implicit learning with explicit retrieval?
\newblock {\em Memory \& cognition}, 21(2):146--155.

\bibitem[Bamberg et~al., 2003]{bamberg2003choice}
Bamberg, S., Ajzen, I., and Schmidt, P. (2003).
\newblock Choice of travel mode in the theory of planned behavior: The roles of past behavior, habit, and reasoned action.
\newblock {\em Basic and applied social psychology}, 25(3):175--187.

\bibitem[Beneduce et~al., 2025]{beneduce2025large}
Beneduce, C., Lepri, B., and Luca, M. (2025).
\newblock Large language models are zero-shot next location predictors.
\newblock {\em IEEE Access}.

\bibitem[Bornstein et~al., 2017]{bornstein2017reminders}
Bornstein, A.~M., Khaw, M.~W., Shohamy, D., and Daw, N.~D. (2017).
\newblock Reminders of past choices bias decisions for reward in humans.
\newblock {\em Nature Communications}, 8(1):15958.

\bibitem[Brown et~al., 2020]{brown2020language}
Brown, T., Mann, B., Ryder, N., Subbiah, M., Kaplan, J.~D., Dhariwal, P., Neelakantan, A., Shyam, P., Sastry, G., Askell, A., et~al. (2020).
\newblock Language models are few-shot learners.
\newblock {\em Advances in neural information processing systems}, 33:1877--1901.

\bibitem[Chen et~al., 2025]{chen2025perceptions}
Chen, R., Wang, C., Sun, Y., Zhao, X., and Xu, S. (2025).
\newblock From perceptions to decisions: Wildfire evacuation decision prediction with behavioral theory-informed llms.
\newblock {\em arXiv preprint arXiv:2502.17701}.

\bibitem[Fulman et~al., 2025]{fulman2025utilizing}
Fulman, N., Memduho{\u{g}}lu, A., and Zipf, A. (2025).
\newblock Utilizing large language models to simulate parking search.
\newblock {\em Transportation Research Part A: Policy and Practice}, 199:104542.

\bibitem[Ge et~al., 2025]{ge2025llm}
Ge, S., Ye, P., Zhang, R., Zhou, M., Dong, H., and Wang, F.-Y. (2025).
\newblock Llm-driven cognitive modeling for personalized travel generation.
\newblock {\em IEEE Transactions on Computational Social Systems}.

\bibitem[Grossmann et~al., 2023]{grossmann2023ai}
Grossmann, I., Feinberg, M., Parker, D.~C., Christakis, N.~A., Tetlock, P.~E., and Cunningham, W.~A. (2023).
\newblock Ai and the transformation of social science research.
\newblock {\em Science}, 380(6650):1108--1109.

\bibitem[Guo et~al., 2025]{guo2025automating}
Guo, X., Yang, X., Peng, M., Lu, H., Zhu, M., and Yang, H. (2025).
\newblock Automating traffic model enhancement with ai research agent.
\newblock {\em Transportation Research Part C: Emerging Technologies}, 178:105187.

\bibitem[Hutson, 2023]{hutson2023can}
Hutson, M. (2023).
\newblock Can ai chatbots replace human subjects in behavioral experiments.
\newblock {\em Science}, 381:6654.

\bibitem[Iida et~al., 1992]{iida1992experimental}
Iida, Y., Akiyama, T., and Uchida, T. (1992).
\newblock Experimental analysis of dynamic route choice behavior.
\newblock {\em Transportation Research Part B: Methodological}, 26(1):17--32.

\bibitem[Jin, 2020]{jin2020stable}
Jin, W.-L. (2020).
\newblock Stable day-to-day dynamics for departure time choice.
\newblock {\em Transportation Science}, 54(1):42--61.

\bibitem[Jotisankasa and Polak, 2006]{jotisankasa2006framework}
Jotisankasa, A. and Polak, J.~W. (2006).
\newblock Framework for travel time learning and behavioral adaptation in route and departure time choice.
\newblock {\em Transportation research record}, 1985(1):231--240.

\bibitem[Kagho et~al., 2020]{kagho2020agent}
Kagho, G.~O., Balac, M., and Axhausen, K.~W. (2020).
\newblock Agent-based models in transport planning: Current state, issues, and expectations.
\newblock {\em Procedia Computer Science}, 170:726--732.

\bibitem[Kojima et~al., 2022]{kojima2022large}
Kojima, T., Gu, S.~S., Reid, M., Matsuo, Y., and Iwasawa, Y. (2022).
\newblock Large language models are zero-shot reasoners.
\newblock {\em Advances in neural information processing systems}, 35:22199--22213.

\bibitem[Lai et~al., 2025]{lai2025llmlight}
Lai, S., Xu, Z., Zhang, W., Liu, H., and Xiong, H. (2025).
\newblock Llmlight: Large language models as traffic signal control agents.
\newblock In {\em Proceedings of the 31st ACM SIGKDD Conference on Knowledge Discovery and Data Mining V. 1}, pages 2335--2346.

\bibitem[Liu et~al., 2023]{liu2023lost}
Liu, N.~F., Lin, K., Hewitt, J., Paranjape, A., Bevilacqua, M., Petroni, F., and Liang, P. (2023).
\newblock Lost in the middle: How language models use long contexts.
\newblock {\em arXiv preprint arXiv:2307.03172}.

\bibitem[Liu et~al., 2025a]{liu2025gatsim}
Liu, Q., Li, C., and Ma, W. (2025a).
\newblock Gatsim: Urban mobility simulation with generative agents.
\newblock {\em arXiv preprint arXiv:2506.23306}.

\bibitem[Liu et~al., 2024]{liu2024can}
Liu, T., Li, M., and Yin, Y. (2024).
\newblock Can large language models capture human travel behavior? evidence and insights on mode choice.
\newblock {\em Avaliable at SSRN 4937575}.

\bibitem[Liu et~al., 2025b]{liu2025aligning}
Liu, T., Li, M., and Yin, Y. (2025b).
\newblock Aligning llm with human travel choices: a persona-based embedding learning approach.
\newblock {\em arXiv preprint arXiv:2505.19003}.

\bibitem[Liu et~al., 2025c]{liu2025llm}
Liu, T., Yang, J., and Yin, Y. (2025c).
\newblock Llm-abm for transportation: Assessing the potential of llm agents in system analysis.
\newblock {\em arXiv preprint arXiv:2503.22718}.

\bibitem[Liu et~al., 2025d]{liu2025toward}
Liu, T., Yang, J., and Yin, Y. (2025d).
\newblock Toward llm-agent-based modeling of transportation systems: A conceptual framework.
\newblock {\em Artificial Intelligence for Transportation}, 1:100001.

\bibitem[Lowry, 2024]{lowry2024multimodal}
Lowry, M.~B. (2024).
\newblock Multimodal experience as a predictor and catalyst of travel behavior.
\newblock {\em Travel Behaviour and Society}, 34:100699.

\bibitem[Mahmassani, 2009]{mahmassani2009learning}
Mahmassani, H.~S. (2009).
\newblock Learning from interactive experiments: Travel behavior and complex.
\newblock In {\em Expanding Sphere of Travel Behaviour Research: Selected Papers from the 11th International Conference on Travel Behaviour Research}, page 131. Emerald Group Publishing.

\bibitem[Meneguzzer and Olivieri, 2013]{meneguzzer2013day}
Meneguzzer, C. and Olivieri, A. (2013).
\newblock Day-to-day traffic dynamics: laboratory-like experiment on route choice and route switching in a simple network with limited feedback information.
\newblock {\em Procedia-Social and Behavioral Sciences}, 87:44--59.

\bibitem[Mo et~al., 2023]{mo2023large}
Mo, B., Xu, H., Zhuang, D., Ma, R., Guo, X., and Zhao, J. (2023).
\newblock Large language models for travel behavior prediction.
\newblock {\em arXiv preprint arXiv:2312.00819}.

\bibitem[Nie et~al., 2025]{nie2025exploring}
Nie, T., Sun, J., and Ma, W. (2025).
\newblock Exploring the roles of large language models in reshaping transportation systems: A survey, framework, and roadmap.
\newblock {\em Artificial Intelligence for Transportation}, 1:100003.

\bibitem[Nishida et~al., 2025]{nishida2025large}
Nishida, R., Ishigaki, T., and Onishi, M. (2025).
\newblock Large language models predict transportation mode choice behavior for a variety of alternative sets.
\newblock {\em Transportation Research Record}, page 03611981251352499.

\bibitem[Qi et~al., 2023]{qi2023investigating}
Qi, H., Jia, N., Qu, X., and He, Z. (2023).
\newblock Investigating day-to-day route choices based on multi-scenario laboratory experiments, part i: Route-dependent attraction and its modeling.
\newblock {\em Transportation Research Part A: Policy and Practice}, 167:103553.

\bibitem[Roccas et~al., 2002]{roccas2002big}
Roccas, S., Sagiv, L., Schwartz, S.~H., and Knafo, A. (2002).
\newblock The big five personality factors and personal values.
\newblock {\em Personality and social psychology bulletin}, 28(6):789--801.

\bibitem[Sameen et~al., 2025]{sameen2025synthesizing}
Sameen, M., Zhang, X., and Zhao, X. (2025).
\newblock Synthesizing attitudes, predicting actions (sapa): Behavioral theory-guided llms for ridesourcing mode choice modeling.
\newblock {\em arXiv preprint arXiv:2509.18181}.

\bibitem[Schlich and Axhausen, 2003]{schlich2003habitual}
Schlich, R. and Axhausen, K.~W. (2003).
\newblock Habitual travel behaviour: evidence from a six-week travel diary.
\newblock {\em Transportation}, 30(1):13--36.

\bibitem[Schwanen and Ettema, 2009]{schwanen2009coping}
Schwanen, T. and Ettema, D. (2009).
\newblock Coping with unreliable transportation when collecting children: examining parents’ behavior with cumulative prospect theory.
\newblock {\em Transportation research part A: Policy and Practice}, 43(5):511--525.

\bibitem[Shohamy and Daw, 2015]{shohamy2015integrating}
Shohamy, D. and Daw, N.~D. (2015).
\newblock Integrating memories to guide decisions.
\newblock {\em Current Opinion in Behavioral Sciences}, 5:85--90.

\bibitem[Song et~al., 2025]{song2025incorporating}
Song, Y.-L., Tsern, C.-E., Wu, C.-C., Chang, Y.-M., Huang, S.-B., Chen, W.-C., Lin, M. C.-L., and Lin, Y.-T. (2025).
\newblock Incorporating llms for large-scale urban complex mobility simulation.
\newblock {\em arXiv preprint arXiv:2505.21880}.

\bibitem[S{\"o}nmez and Graefe, 1998]{sonmez1998determining}
S{\"o}nmez, S.~F. and Graefe, A.~R. (1998).
\newblock Determining future travel behavior from past travel experience and perceptions of risk and safety.
\newblock {\em Journal of travel research}, 37(2):171--177.

\bibitem[Team et~al., 2023]{team2023gemini}
Team, G., Anil, R., Borgeaud, S., Alayrac, J.-B., Yu, J., Soricut, R., Schalkwyk, J., Dai, A.~M., Hauth, A., Millican, K., et~al. (2023).
\newblock Gemini: a family of highly capable multimodal models.
\newblock {\em arXiv preprint arXiv:2312.11805}.

\bibitem[Touvron et~al., 2023]{touvron2023llama}
Touvron, H., Lavril, T., Izacard, G., Martinet, X., Lachaux, M.-A., Lacroix, T., Rozi{\`e}re, B., Goyal, N., Hambro, E., Azhar, F., et~al. (2023).
\newblock Llama: Open and efficient foundation language models.
\newblock {\em arXiv preprint arXiv:2302.13971}.

\bibitem[Tzachristas et~al., 2025]{tzachristas2025guided}
Tzachristas, I., Narayanan, S., and Antoniou, C. (2025).
\newblock Guided persona-based ai surveys: Can we replicate personal mobility preferences at scale using llms?
\newblock {\em arXiv preprint arXiv:2501.13955}.

\bibitem[Wang et~al., 2024]{wang2024large}
Wang, J., Jiang, R., Yang, C., Wu, Z., Onizuka, M., Shibasaki, R., Koshizuka, N., and Xiao, C. (2024).
\newblock Large language models as urban residents: An llm agent framework for personal mobility generation.
\newblock {\em Advances in Neural Information Processing Systems}, 37:124547--124574.

\bibitem[Wang et~al., 2025a]{wang2025agentic}
Wang, L., Duan, P., He, Z., Lyu, C., Chen, X., Zheng, N., Yao, L., and Ma, Z. (2025a).
\newblock Agentic large language models for day-to-day route choices.
\newblock {\em Transportation Research Part C: Emerging Technologies}, 180:105307.

\bibitem[Wang et~al., 2025b]{wang2025comparing}
Wang, L., Jiang, Z., Hu, C., Zhao, J., Zhu, Z., Chen, X., Wang, Z., Liu, T., He, G., Yin, Y., et~al. (2025b).
\newblock Comparing ai and human decision-making mechanisms in daily collaborative experiments.
\newblock {\em iScience}, 28(6).

\bibitem[Wang et~al., 2023]{wang2023would}
Wang, X., Fang, M., Zeng, Z., and Cheng, T. (2023).
\newblock Where would i go next? large language models as human mobility predictors.
\newblock {\em arXiv preprint arXiv:2308.15197}.

\bibitem[Wei et~al., 2021]{wei2021finetuned}
Wei, J., Bosma, M., Zhao, V.~Y., Guu, K., Yu, A.~W., Lester, B., Du, N., Dai, A.~M., and Le, Q.~V. (2021).
\newblock Finetuned language models are zero-shot learners.
\newblock {\em arXiv preprint arXiv:2109.01652}.

\bibitem[Xu and Wang, 2025]{xu2025can}
Xu, P. and Wang, D. (2025).
\newblock Can large language models trigger a paradigm shift in travel behavior modeling? experiences with modeling travel satisfaction.
\newblock {\em arXiv preprint arXiv:2505.23262}.

\bibitem[Xu and Jiao, 2025]{xu2025evaluating}
Xu, Y. and Jiao, J. (2025).
\newblock Evaluating retrieval-augmented generation strategies for large language models in travel mode choice prediction.
\newblock {\em arXiv preprint arXiv:2508.17527}.

\bibitem[Yan et~al., 2025a]{yan2025addressing}
Yan, X., Dai, T., et~al. (2025a).
\newblock Addressing the alignment problem in transportation policy making: an llm approach.
\newblock {\em arXiv preprint arXiv:2510.13139}.

\bibitem[Yan et~al., 2025b]{yan2025valuing}
Yan, Y., Liu, T., and Yin, Y. (2025b).
\newblock Valuing time in silicon: Can large language model replicate human value of travel time.
\newblock {\em arXiv preprint arXiv:2507.22244}.

\bibitem[Yu et~al., 2020]{yu2020day}
Yu, Y., Han, K., and Ochieng, W. (2020).
\newblock Day-to-day dynamic traffic assignment with imperfect information, bounded rationality and information sharing.
\newblock {\em Transportation Research Part C: Emerging Technologies}, 114:59--83.

\bibitem[Yuksekgonul et~al., 2025]{yuksekgonul2025optimizing}
Yuksekgonul, M., Bianchi, F., Boen, J., Liu, S., Lu, P., Huang, Z., Guestrin, C., and Zou, J. (2025).
\newblock Optimizing generative ai by backpropagating language model feedback.
\newblock {\em Nature}, 639(8055):609--616.

\bibitem[Zhang et~al., 2018]{zhang2018cumulative}
Zhang, C., Liu, T.-L., Huang, H.-J., and Chen, J. (2018).
\newblock A cumulative prospect theory approach to commuters’ day-to-day route-choice modeling with friends’ travel information.
\newblock {\em Transportation Research Part C: Emerging Technologies}, 86:527--548.

\bibitem[Zhang and Xu, 2025]{zhang2025transmode}
Zhang, M. and Xu, Y. (2025).
\newblock Transmode-llm: Feature-informed natural language modeling with domain-enhanced prompting for travel behavior modeling.
\newblock In {\em Large Language Models for Scientific and Societal Advances}.

\bibitem[Zhao and Huang, 2016]{zhao2016experiment}
Zhao, C.-L. and Huang, H.-J. (2016).
\newblock Experiment of boundedly rational route choice behavior and the model under satisficing rule.
\newblock {\em Transportation Research Part C: Emerging Technologies}, 68:22--37.

\bibitem[Zhou and Mahmassani, 2007]{zhou2007structural}
Zhou, X. and Mahmassani, H.~S. (2007).
\newblock A structural state space model for real-time traffic origin--destination demand estimation and prediction in a day-to-day learning framework.
\newblock {\em Transportation Research Part B: Methodological}, 41(8):823--840.

\end{thebibliography}
